\documentclass{article} 
\usepackage{iclr2027_conference}

\usepackage{amsmath,amsfonts,bm}

\def\eqref#1{equation~\ref{#1}}

\def\1{\bm{1}}

\DeclareMathAlphabet{\mathsfit}{\encodingdefault}{\sfdefault}{m}{sl}
\SetMathAlphabet{\mathsfit}{bold}{\encodingdefault}{\sfdefault}{bx}{n}

\usepackage{newtxtext,newtxmath}  
\usepackage{hyperref}
\usepackage{url}
\usepackage{booktabs}
\usepackage{multirow}
\usepackage{amsmath}
\usepackage{graphicx}
\usepackage{colortbl}
\usepackage{tabularx}
\usepackage{array}
\definecolor{sota}{rgb}{0.894,0.886,0.961} 
\definecolor{tableoneblue}{HTML}{D9EEFF} 
\definecolor{grouphdr}{HTML}{E6E6E6} 
\usepackage{subcaption}
\usepackage{float}
\usepackage[belowfloats]{footmisc}
\usepackage{microtype}
\usepackage{enumitem}
\usepackage{listings}
\lstdefinestyle{jsonout}{basicstyle=\ttfamily\scriptsize,breaklines=true,breakatwhitespace=false,columns=fullflexible,keepspaces=true,showstringspaces=false,aboveskip=0pt,belowskip=0pt}
\lstdefinestyle{rawtext}{style=jsonout,resetmargins=true,breakindent=0pt,breakautoindent=false,prebreak={},postbreak={}}
\usepackage[most]{tcolorbox}

\title{SAG: SQL-Retrieval Augmented Generation with Query-Time Dynamic Hyperedges}
\author{Yuchao Wu\thanks{Correspondence to: \texttt{jomy@zleap.com}.}, Junqin Li, XingCheng Liang, Yongjie Chen, Yinghao Liang \\
{\bf Linyuan Mo, Guanxian Li} \\
Zleap AI \\
\texttt{\{jomy,junqing,lensen,jinzhoulawen,leo\}@zleap.com} \\
\texttt{\{mo-linyuan,li\_guanxian\}@foxmail.com}
}

\iclrfinalcopy 

\begin{document}

\maketitle
\lhead{}

\begin{abstract}
While retrieval-augmented generation (RAG) has proven effective at giving LLMs access to external knowledge, mainstream dense-retrieval implementations remain inherently limited in handling structured constraints and multi-hop reasoning. Graph-based methods address this by constructing knowledge graphs offline, but they often fragment semantics, incur high maintenance, and complicate incremental updates. We propose SAG (SQL-Retrieval Augmented Generation), a structured retrieval architecture that organizes documents into an event-entity index without building a global knowledge graph. SAG represents each chunk as a semantically complete event paired with its entities, forming a latent hyperedge that preserves n-ary relations without decomposing them into triples. At query time, SAG treats shared entities as join keys to connect related chunks. This dynamically yields a query-scoped neighborhood of events, and yet every piece of evidence remains the original chunk throughout. Experiments on HotpotQA, 2WikiMultiHopQA, and MuSiQue show that SAG achieves the best retrieval and end-to-end QA performance on every benchmark, with gains that widen as reasoning-chain complexity increases. On MuSiQue, where multi-hop evidence chaining is most demanding, SAG reaches 80.36\% Recall@5, outperforming the strongest baseline by 11.52 points. This work paves the way for knowledge infrastructure that enables LLM agents to retrieve and reason over continually growing organizational knowledge.\footnote{Benchmark and evaluation code:
\url{https://github.com/Zleap-AI/SAG-Benchmark}}
\end{abstract}

\section{Introduction}
\label{sec:intro}

What an agent can do is bounded by what it can retrieve. Organizational knowledge is distributed across documents, messages, customer systems, and code; much of it is unstructured, relational, and continuously accumulating. Making this knowledge available to agents is therefore an infrastructure problem as well as a modeling problem. Multi-hop question answering makes this gap concrete. Answers often require evidence from multiple documents, whereas conventional RAG pipelines~\citep{lewis2020rag,karpukhin2020dpr} rank passages independently by their similarity to the query. This approach works well when a single passage contains the answer, but it does not explicitly represent associations across passages. In multi-hop retrieval, failing to recover one intermediate passage can break the entire evidence chain~\citep{yang2018hotpotqa,trivedi2022musique,mavi2024multihop}. The challenge is therefore not only to improve single-shot recall, but also to organize evidence so that cross-document associations can be recovered reliably.

Structure-augmented methods address this problem by constructing knowledge graphs or hierarchical representations offline~\citep{edge2024graphrag,gutierrez2025hipporag2}. When n-ary events are decomposed into independently indexed triples, however, their shared event boundary may be lost. Hypergraph-based methods~\citep{feng2025hypergraphrag} preserve higher-order relations but still encode them in a persistent corpus-level structure. These structures support graph traversal and ranking, but the connections available during retrieval are determined before the query arrives, while ingestion and maintenance remain separate from query execution. This raises a different question: can the structure needed by a query be activated from an append-only index during retrieval, without constructing a global knowledge graph?

\begin{figure}[!b]
\centering
\includegraphics[width=\linewidth]{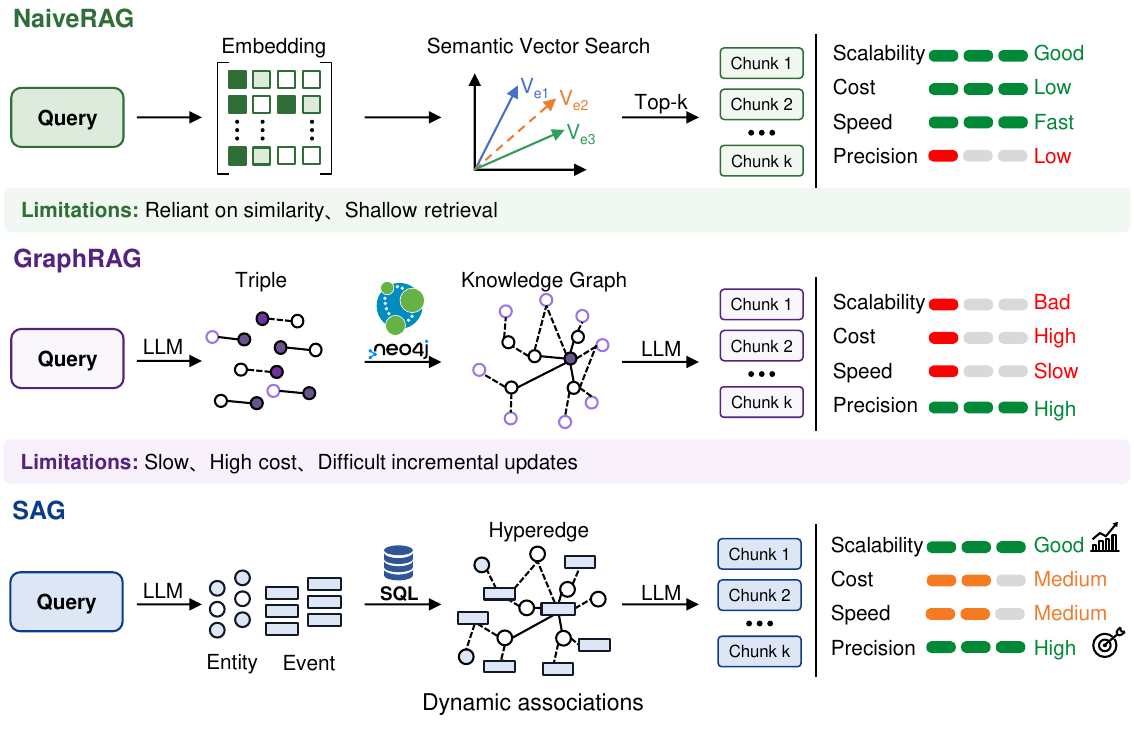}
\caption{Three RAG paradigms. NaiveRAG retrieves top-$k$ chunks by dense similarity, while GraphRAG organizes evidence in an offline knowledge graph. SAG stores event--entity relations in an append-only index and activates query-specific latent hyperedges through SQL joins.}
\label{fig:paradigm_comparison}
\end{figure}

Our central claim is that multi-hop retrieval need not rely exclusively on either dense similarity or a globally constructed graph (Figure~\ref{fig:paradigm_comparison}). The required associative structure is latent in the events described by passages and the entities shared across them. SAG exposes this structure through an event--entity index. Each chunk is independently represented as one semantically complete event linked to a set of indexing entities. Because an event binds all its entities within one retrieval unit, it defines a latent hyperedge that retains the event's n-ary relation rather than decomposing it into independent triples. At query time, SQL joins over shared entities activate a local neighborhood of related events. This structural path is combined with dense retrieval, after which an LLM selects evidence from the compressed candidate set. SAG constructs no global corpus graph, and selected events are mapped back to their original chunks before answer generation. Under unified evaluation settings, SAG achieves the best retrieval and end-to-end QA performance on HotpotQA, 2WikiMultiHopQA, and MuSiQue. Its largest retrieval gain occurs on MuSiQue, where SAG reaches 80.36\% Recall@5 and exceeds the strongest baseline by 11.52 points.

To summarize, our contributions are fourfold. (1) We introduce SAG, a retrieval architecture that activates query-specific latent hyperedges over an event--entity index through SQL joins, without constructing a global knowledge graph. (2) We formalize the event--entity representation, characterize the neighborhoods recovered through incidence joins, and compare its corpus-level connectivity with four structure-augmented baselines. (3) We evaluate SAG on three multi-hop benchmarks under a unified configuration and isolate the effects of representation, expansion, final selection, and candidate budget. (4) We evaluate robustness to embedding replacement and continual corpus growth, and show how bounded candidate activation limits the downstream SQL and LLM workloads. Beyond the benchmark study, we provide an engineering implementation inspired by SAG's core mechanisms.\footnote{Engineering project: \url{https://github.com/Zleap-AI/SAG}} Together, these properties make SAG a step toward knowledge infrastructure for agents operating over continually growing corpora.

\section{Related Work}
\label{sec:related}

\textbf{Retrieval-Augmented Generation.}
RAG combines a parametric generator with a dense passage index to incorporate external knowledge~\citep{lewis2020rag}. Later work adapts when and how retrieval is performed. Self-RAG uses reflection tokens to decide when to retrieve and to evaluate retrieved evidence~\citep{asai2024selfrag}, while FLARE retrieves from predicted upcoming sentences when generation contains low-confidence tokens~\citep{jiang2023flare}. Adaptive-RAG routes queries among no-retrieval, single-step, and iterative strategies according to predicted complexity~\citep{jeong2024adaptiverag}. IRCoT instead alternates retrieval with chain-of-thought generation, using each reasoning step to guide the next retrieval~\citep{trivedi2023ircot}. These methods modify retrieval control or query formulation, whereas SAG changes how corpus knowledge is represented and connected.

\textbf{Structure-Augmented Retrieval.}
GraphRAG constructs an entity graph and community summaries for answering global corpus-level questions~\citep{edge2024graphrag}. HippoRAG builds a knowledge graph from extracted relations and propagates relevance with personalized PageRank, while HippoRAG~2 integrates passages more directly and strengthens the online use of an LLM~\citep{gutierrez2024hipporag1,gutierrez2025hipporag2}. LightRAG combines incremental graph updates with low- and high-level retrieval over entities and relations~\citep{guo2025lightrag}. SiReRAG constructs similarity and relatedness trees and flattens their nodes into a unified retrieval pool~\citep{zhang2025sirerag}, whereas StructRAG selects and constructs a task-appropriate structure at inference time~\citep{li2025structrag}. These methods either maintain corpus-level structures or reconstruct structured views for individual queries. SAG instead stores event--entity incidence records and activates query-specific associations through SQL joins without materializing a global graph. ChatDB and StructGPT query existing databases or structured sources, but do not organize unstructured corpora for retrieval~\citep{hu2023chatdb,jiang2023structgpt}.

\textbf{Hyperedge and Higher-Order Representations.}
Hypergraphs and n-ary knowledge-base models represent relations among multiple entities without reducing each relation to independent binary triples~\citep{zhou2006hypergraph,feng2019hgnn,fatemi2019hype,galkin2020stare,liu2020getd,liu2021ram}. HyperGraphRAG extracts n-ary facts and stores a corpus-level hypergraph with vector indexes for entities and hyperedges~\citep{feng2025hypergraphrag}. Hyper-RAG retrieves both entities and correlations from a hypergraph repository and expands them through structural diffusion~\citep{feng2026hyperrag}. HGRAG treats entities as nodes and passages as hyperedges, combining entity- and passage-level similarity through hypergraph diffusion~\citep{wang2026hgrag}. Graph-R1 constructs a lightweight knowledge hypergraph and learns multi-turn retrieval through reinforcement learning~\citep{luo2026graphr1}. These methods materialize a corpus-level hypergraph. SAG instead stores each event and its entities as relational incidence rows. SQL joins over shared entities activate only the latent hyperedges required by the query, after which the original chunks are returned to the reader.

\section{SAG}

\begin{figure}[t]
    \centering
    \includegraphics[width=0.9\linewidth]{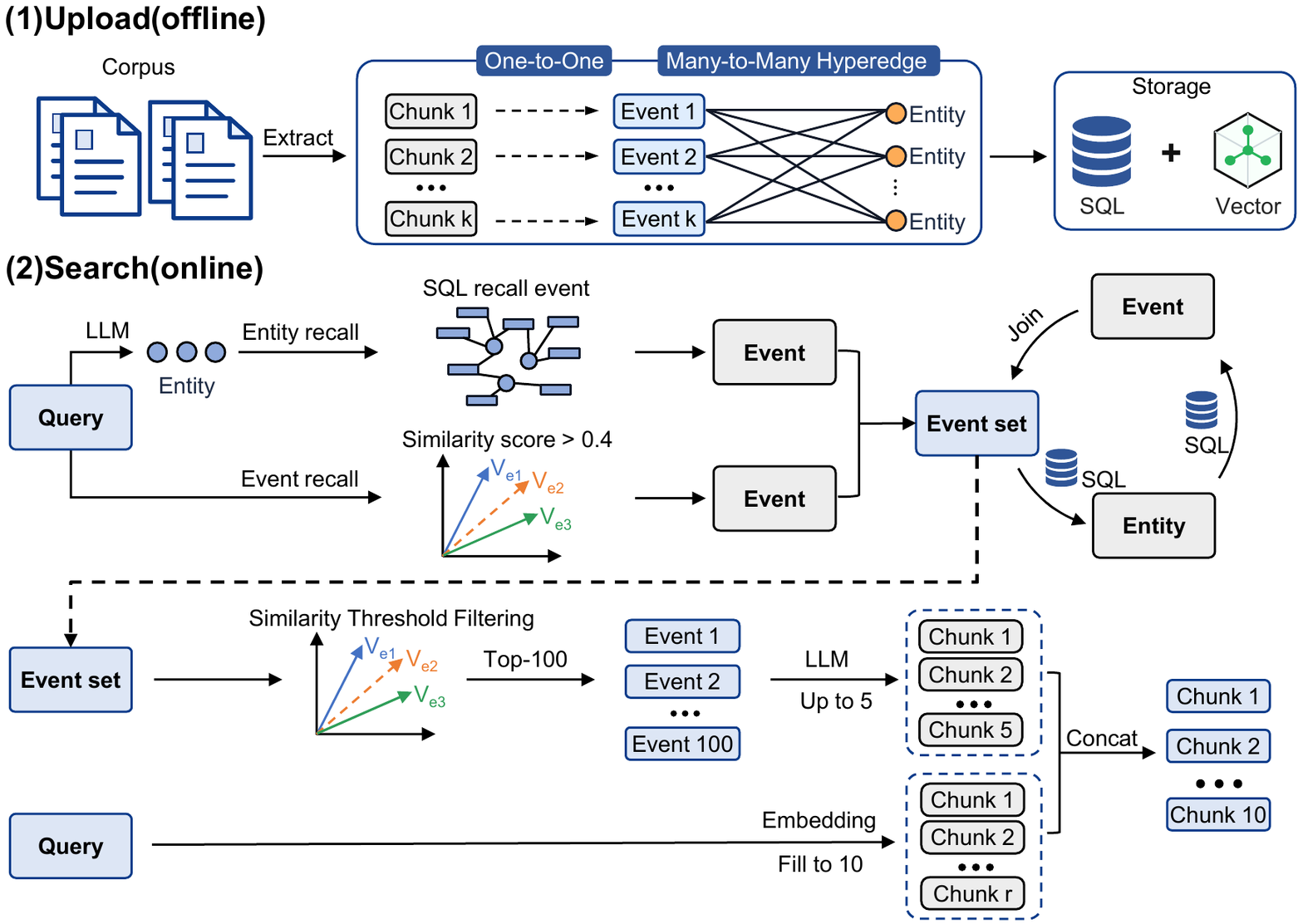}
    \caption{Architecture overview of SAG. Offline, chunks are indexed as
    events and entities across SQL, vector, and full-text stores; online,
    initial recall is expanded at query time, and final selection operates
    over the compressed candidate set. The output panel illustrates the case
    in which the structural path reaches its five-event cap; the semantic path
    fills the remaining slots to return ten chunks.}
    \label{fig:architecture}
\end{figure}

SAG has an offline indexing phase and an online phase with seed retrieval,
query-time expansion, and final selection (Figure~\ref{fig:architecture}). SQL performs exact entity joins, vector
retrieval finds semantically relevant seed chunks even when the query and
the passage share no exact terms, and the LLM is reserved for
event-entity extraction, query-entity identification, and final selection
over the compressed candidate set.

\subsection{Event-Entity Index}
\label{subsec:index}

SAG builds an event-entity index rather than a global knowledge graph. Each
chunk is processed independently into one event $e$ and a set of entities
$V(e)$, which together define a latent hyperedge. New documents can therefore
be appended without recomputing existing records.

\textbf{Event.}
An event is a concise statement of the chunk's core content, with one event
extracted per chunk. Decomposing a chunk into independent triples can fragment
its meaning, especially when several entities jointly constrain the underlying
fact. SAG instead keeps the event intact: retrieval operates on the concise
event text, while the corresponding original chunk is returned as evidence.

\textbf{Entities.}
Entities span 11 semantic types.\footnote{The complete inventory is time,
location, person, organization, group, subject, work, product, action, metric,
and tags; see Appendix~\ref{appendix:indexing-prompt}.} They are not treated as self-contained evidence; instead, they serve
primarily as indexing and expansion keys that connect events sharing the same
participant or concept. Events and entities are produced in parallel by a
single LLM call per chunk and written into SQL as many-to-many rows, as well as
into the vector and full-text indices.

The index deliberately avoids full entity disambiguation. It relies only on
string normalization and SQL deduplication, preserving independent chunk
processing and append-only ingestion. This tolerance for imperfect
normalization is a deployment choice: production corpora arrive continuously and
are rarely clean, and requiring global disambiguation or re-chunking before
indexing would forfeit the append-only property that keeps ingestion practical
as the corpus grows.

\subsection{Latent Hyperedges}
\label{subsec:hyperedge}

Let $V$ denote the entity vertices and $H$ the indexed events (hyperedges). Each event
$h \in H$ defines a hyperedge over its entity set $V(h) \subseteq V$, keeping
all entities mentioned in the event together as one retrieval unit. The event
text preserves the relational meaning that a bare set of entity names cannot
express, while the linked original chunk remains the evidence returned to the
downstream reader.

The event-entity table stores each hyperedge as a set of rows $(h,v)$ for
$v \in V(h)$. Grouping these rows by event identifier recovers the complete
entity set of the event, making the representation lossless with respect to
incidence.

One expansion round (Section~\ref{subsec:expansion}) walks from seed events to
their entities and then back to co-incident events. The activated structure
exists only for the current query, and ingesting a new chunk requires only
appending new rows rather than rebuilding a global graph. Corpus statistics
support this structural interpretation, with SAG showing less fragmentation
than GraphRAG, LightRAG, and HyperGraphRAG on MuSiQue and more than 93\%
giant-component coverage on every benchmark (Appendix~\ref{appendix:connectivity}).
Section~\ref{sec:Analysis} further shows that hyperedges outperform triple
decomposition in Recall@5.

\subsection{Seed Retrieval}

Given a query $q$, SAG first defines a query-specific working corpus $D_q$. In
the full-pool configuration used for the main results, $D_q = D$. Under bounded
activation, embedding retrieval first selects the top $K_{\mathrm{pool}}$
chunks from the global corpus, and all subsequent event and entity operations
are restricted to records associated with those chunks. This corpus-level
activation budget is distinct from $K_{\mathrm{cand}}$, which controls the
number of events passed to the final LLM reranker.

Within $D_q$, SAG constructs an initial event set $R$ through two parallel
retrieval paths, each addressing a failure mode of the other.

\textbf{Path A: entity-guided structured recall.}
A lightweight LLM call identifies key entities in the query and produces a
seed set $Q = \{v_i\}$. Similarity search over the entity vector index expands
these seeds into an augmented set $\widehat{Q} \supseteq Q$ using
$\operatorname{Sim}_{\mathrm{ent}}: V \times V \to [0,1]$.
Because entities
are short strings, SAG uses a high similarity threshold $\tau_{\mathrm{ent}}=0.9$ by default;
looser matching is more likely to conflate distinct entities than to recover
genuine aliases. A SQL join then retrieves all events sharing at least one
entity with $\widehat{Q}$:
\begin{equation}
R_{\mathrm{ent}}
=
\left\{
h \in H
\;\middle|\;
V(h) \cap \widehat{Q} \neq \varnothing
\right\}.
\end{equation}

\textbf{Path B: direct event recall via query vector.}
In parallel, SAG retrieves events directly by embedding similarity to the
query using $\operatorname{Sim}_{\mathrm{evt}}: H \times \{q\} \to [0,1]$.
Because events are longer semantic statements, this path uses a
deliberately low threshold $\tau_{\mathrm{evt}}=0.4$ by default, to favor recall and
leave precision control to later ranking. The retrieved events form
$R_{\mathrm{evt}}$.

The two paths are merged to form the seed event set:
\begin{equation}
R
=
R_{\mathrm{ent}} \cup R_{\mathrm{evt}}.
\end{equation}

\subsection{Expansion and Selection}
\label{subsec:expansion}

\textbf{Expansion.}
Seed retrieval finds events that either share entities with the query or have
high direct semantic similarity. Multi-hop questions, however, may require
evidence connected through intermediate entities that do not appear in the
query.

Starting from $R$, SAG uses a reverse SQL join to collect the entities
associated with the seed events. Previously unexplored entities form an
expansion frontier, which is pruned to a fixed budget before the next join to
bound both computation and candidate growth. SAG then retrieves previously
unseen events associated with the retained frontier entities.

Each event-to-entity-to-event composition constitutes one expansion round.
Expansion runs for at most $L$ rounds, with $L=1$ by default, and each round
introduces only entities and events not visited earlier. Let $X \subseteq H$ denote the
events added through expansion. The complete candidate pool is
\begin{equation}
C
=
R \cup X,
\end{equation}
\textbf{Coarse ranking.}
SAG ranks the events in $C$ by embedding similarity to the query and retains
the top $K_{\mathrm{cand}}$ events:
\begin{equation}
\widehat{C}
=
\operatorname{TopK}_{K_{\mathrm{cand}}}(C,q).
\end{equation}

\textbf{Dual-path output.}
The structural path preserves evidence recovered through cross-entity
expansion, which may have low direct query similarity despite being necessary
for multi-step reasoning. The semantic path retains chunks that are directly
relevant to the query but may not participate in the expanded structure.

The system returns $K_{\mathrm{out}}=10$ chunks. In the structural path, the
LLM reads $\widehat{C}$ in context and selects up to
$K_{\mathrm{event}}=5$ events that jointly entail the answer, producing
$E_{\mathrm{sel}}$ with $|E_{\mathrm{sel}}|\leq K_{\mathrm{event}}$. These
events are mapped to chunks as
$D_{\mathrm{evt}}=\{\phi(h):h\in E_{\mathrm{sel}}\}$ through the event-to-chunk
mapping $\phi:H\to D$. This is a contextual selection step rather than
a pointwise reranking: cross-candidate attention lets the model detect the
implicit relations between events and return the chain sufficient to support an
answer, which a per-event score cannot capture. The semantic path produces a
ranked chunk list $R_{\mathrm{chunk}}$ by direct retrieval over the chunk index.
After removing chunks already in $D_{\mathrm{evt}}$, it fills the remaining
$K_{\mathrm{direct}}(q)=K_{\mathrm{out}}-|D_{\mathrm{evt}}|$ slots. Thus,
$K_{\mathrm{event}}$ is a cap on structural selections rather than a fixed
five-chunk allocation, and $K_{\mathrm{direct}}$ varies by query:
\begin{equation}
D_{\mathrm{out}}
=
D_{\mathrm{evt}}
\cup
\operatorname{TopK}_{K_{\mathrm{direct}}(q)}
\left(R_{\mathrm{chunk}}\setminus D_{\mathrm{evt}}\right).
\end{equation}

These explicit stages make retrieval failures traceable.
Section~\ref{sec:Analysis} shows that most expansion gains arrive in the first
round, contextual LLM selection outperforms pointwise reranking, and the two
output paths are complementary.
Appendix~\ref{appendix:trace} presents successful and unsuccessful activation traces.

\section{Experiments}

\subsection{Datasets and Metrics}

We evaluate on three multi-hop benchmarks: MuSiQue~\citep{trivedi2022musique} (answerable subset, up to four counterfactually filtered hops), 2WikiMultiHopQA~\citep{ho2020wikimultihop} (inference-type multi-document questions), and HotpotQA~\citep{yang2018hotpotqa} (two-hop). Following the HippoRAG~2 evaluation setup~\citep{gutierrez2025hipporag2}, which builds on the IRCoT protocol~\citep{trivedi2023ircot}, we sample 1,000 development questions per multi-hop dataset and use the same pooled retrieval corpora of supporting and distractor passages (11,656 passages for MuSiQue, 6,119 for 2WikiMultiHopQA, and 9,811 for HotpotQA). For corpus-growth analysis, we also follow its NQ setup~\citep{gutierrez2025hipporag2}: 1,000 sampled queries and 9,633 passages, with the NQ corpus derived from~\citet{wang2024rear}.

We report Recall@5 in the main results and Recall@1/2/5/10 in Appendix~\ref{appendix:full-results}. For QA we report F1 and four LLM-judge metrics following \citet{xiang2025use}: answer correctness, coverage, context relevancy, and evidence recall (definitions in Appendix~\ref{appendix:llm-judge}).

\subsection{Baselines and Implementation}
\label{subsec:baselines}

We compare SAG against three families of baselines. Simple retrievers include BM25 and Contriever. Large embedding models include BGE-Large-EN-v1.5~\citep{xiao2023bge}, GTE-Qwen2-7B~\citep{li2023gte}, GritLM-7B~\citep{muennighoff2025gritlm}, and NV-Embed-v2~\citep{lee2024nvembed}. Structure-augmented methods include GraphRAG~\citep{edge2024graphrag}, LightRAG~\citep{guo2025lightrag}, HippoRAG~2~\citep{gutierrez2025hipporag2}, HyperGraphRAG~\citep{feng2025hypergraphrag}, and HyperRAG~\citep{feng2026hyperrag}. For all structure-augmented methods we use BGE-Large-EN-v1.5 as the retriever and Qwen3.6-Flash~\citep{qwenteam2025qwen3} as the reader and, where applicable, the extractor and reranker. This controlled setup isolates retrieval architecture from model choice.

SAG stores its index in MySQL and Elasticsearch, with dense-vector search handled by Elasticsearch rather than a separate vector database. Unless otherwise noted, we set expansion depth $L=1$, seed budget $K_{\mathrm{seed}}=50$, frontier budget 50, and candidate budget $K_{\mathrm{cand}}=100$. SAG returns $K_{\mathrm{out}}=10$ chunks per query: at most $K_{\mathrm{event}}=5$ from structural selection, with the remainder filled by direct retrieval.

\section{Results and Analysis}
\label{sec:results}

The evaluation is organized around three questions. First, holding the embedding model and the reader fixed across all methods, does the event-entity structure that SAG adds improve retrieval, and does that improvement carry through to end-to-end answers. Second, which stage of the pipeline accounts for the gain, and can that gain instead be matched by a stronger off-the-shelf component. Third, does the advantage persist under realistic operating conditions, namely a corpus that keeps growing and a per-query budget that must stay bounded. We take them in order.

\subsection{Main Results}
\label{sec:mainresults}

\begin{table}[t]
\centering
\small
\caption{Main results under a unified evaluation setup. All methods use Qwen3.6-Flash as the reader, and all structure-augmented methods use BGE-Large-EN-v1.5 as the retriever. Avg is the mean over the three benchmarks. Recall@5 is marked -- where a method returns generated answers rather than ranked passages (all structure-augmented baselines except HippoRAG~2). Full breakdowns are in Appendix~\ref{appendix:full-results} and~\ref{appendix:full-qa}. \textbf{Best} and \underline{second-best} results are highlighted.}
\label{table-main-results}
\setlength{\tabcolsep}{4pt}
\begin{tabularx}{\linewidth}{l*{8}{>{\centering\arraybackslash}X}}
\toprule
Method & \multicolumn{2}{c}{MuSiQue} & \multicolumn{2}{c}{2Wiki} & \multicolumn{2}{c}{HotpotQA} & \multicolumn{2}{c}{Avg} \\
\cmidrule(lr){2-3}\cmidrule(lr){4-5}\cmidrule(lr){6-7}\cmidrule(lr){8-9}
& R@5 & F1 & R@5 & F1 & R@5 & F1 & R@5 & F1 \\
\midrule
\multicolumn{9}{c}{\cellcolor{grouphdr}\textit{Simple Baselines}} \\
BM25~\citep{robertson1994bm25} & 42.11 & 35.19 & 63.78 & 56.26 & 73.05 & 64.64 & 59.65 & 52.03 \\
Contriever~\citep{izacard2022contriever} & 45.32 & 34.39 & 59.25 & 50.62 & 74.85 & 64.06 & 59.81 & 49.69 \\
\multicolumn{9}{c}{\cellcolor{grouphdr}\textit{Large Embedding Models}} \\
BGE-Large-EN-v1.5~\citep{xiao2023bge} & 56.32 & 42.80 & 69.83 & 61.49 & 89.20 & 74.81 & 71.78 & 59.70 \\
GTE-Qwen2-7B~\citep{li2023gte} & 61.92 & 44.60 & 74.18 & 64.91 & 88.45 & 73.04 & 74.85 & 60.85 \\
GritLM-7B~\citep{muennighoff2025gritlm} & 66.22 & 49.30 & 75.62 & 66.12 & 93.40 & 73.22 & 78.41 & 62.88 \\
NV-Embed-v2~\citep{lee2024nvembed} & \underline{68.84} & 52.88 & 77.00 & 68.45 & \underline{95.25} & \underline{78.00} & 80.36 & 66.44 \\
\multicolumn{9}{c}{\cellcolor{grouphdr}\textit{Structure-Augmented RAG}} \\
GraphRAG~\citep{edge2024graphrag} & -- & \underline{54.14} & -- & 72.08 & -- & 76.29 & -- & 67.50 \\
LightRAG~\citep{guo2025lightrag} & -- & 46.66 & -- & 65.75 & -- & 73.09 & -- & 61.83 \\
HyperGraphRAG~\citep{feng2025hypergraphrag} & -- & 49.20 & -- & 63.44 & -- & 74.70 & -- & 62.45 \\
HyperRAG~\citep{feng2026hyperrag} & -- & 51.80 & -- & \underline{76.89} & -- & 77.19 & -- & \underline{68.63} \\
HippoRAG~2~\citep{gutierrez2025hipporag2} & 65.13 & 47.49 & \underline{90.35} & 74.34 & 94.35 & 77.24 & \underline{83.28} & 66.36 \\
\rowcolor{tableoneblue} SAG (Ours) & \textbf{80.36} & \textbf{61.15} & \textbf{93.34} & \textbf{78.06} & \textbf{96.50} & \textbf{79.66} & \textbf{90.07} & \textbf{72.96} \\
\bottomrule
\end{tabularx}
\end{table}

\textbf{Retrieval improves most where reasoning chains are longest.} Table~\ref{table-main-results} reports Recall@5 under a unified configuration in which every structure-augmented method shares the BGE-Large-EN-v1.5 retriever and the Qwen3.6-Flash reader, so architecture is the only free variable. SAG attains the best retrieval result on all three benchmarks, and the margin grows with chain depth. On MuSiQue, whose counterfactual filtering removes single-hop shortcuts and whose questions require up to four non-skippable hops, SAG reaches 80.36\% Recall@5 against 65.13\% for HippoRAG~2, a 15.23-point lead. On two-hop HotpotQA the Recall@5 margin narrows to 2.15 points, yet SAG still leads HippoRAG~2 in Recall@2 by 13.20 points. This gradient is what the expansion mechanism predicts. SQL joins add connected events without multiplicative decay, so evidence separated from the seed by several join steps still reaches the candidate set, whereas HippoRAG~2 propagates embedding-derived scores through personalized PageRank, whose damping attenuates exactly the long-range connections that four-hop questions depend on. 

The 2WikiMultiHopQA result is worth a note because it exposed a flaw in our initial design. SAG initially deferred pruning until final vector ranking, allowing high-similarity but irrelevant expansions to displace lower-similarity evidence and limiting Recall@5 to 88.00\%, below HippoRAG~2's 90.35\%. Its traceable intermediate states (Section~\ref{subsec:expansion}) revealed this failure. Pruning events and entities before and during expansion raises Recall@5 to 93.34\%, surpassing HippoRAG~2.

\textbf{Higher recall translates into better answers.}
Retrieval gains matter only when they improve the final answer. With the reader fixed across all methods, SAG achieves the highest F1 on each benchmark in Table~\ref{table-main-results}. The advantage is largest on MuSiQue, where SAG reaches 61.15 F1 and exceeds GraphRAG by 7.01 points. On 2WikiMultiHopQA and HotpotQA, SAG leads the strongest baseline by 1.17 and 1.66 points, respectively. This pattern closely follows the retrieval results. The larger gain on MuSiQue indicates that SAG's retrieval advantage becomes more consequential when answering depends on longer evidence chains. The additional passages recovered by SAG therefore contribute to answer generation rather than merely increasing recall with loosely related context.

\textbf{LLM-based evaluation supports the main results.}
Using Qwen3.7-Plus, SAG achieves the best answer correctness, coverage, and evidence recall on all three benchmarks (Table~\ref{table-llm-judge}). The only exception is context relevancy, which should be interpreted alongside the judge-input configuration. Several baselines provide richer retrieval outputs, whereas SAG and HippoRAG~2 are evaluated using only their top-5 original chunks (Table~\ref{table-qa-context}). Because this metric assesses whether the supplied context is sufficient to answer the question, broader inputs may have an advantage. Even with its narrower input, SAG remains within 1.63--5.97 points of the best context-relevancy scores. This pattern is consistent with SAG's architecture. Shared-entity expansion finds bridge evidence, and contextual selection retains passages that together form a complete evidence chain. SAG then returns the original chunks, preserving their full context.

\subsection{Analysis} \label{sec:Analysis}

\textbf{Ablation study.}
We conduct the ablation on MuSiQue, whose counterfactual filtering makes differences in evidence chaining easier to observe. Table~\ref{table-ablation} shows that final selection has the largest measured effect. Replacing Qwen3.6-Flash with Qwen3-Reranker-8B reduces Recall@5 by 13.25 points to 67.11\%. The reranker scores candidates independently, whereas Qwen3.6-Flash evaluates the candidate set jointly and can account for complementary evidence. Even with the reranker, SAG remains above HippoRAG~2 at 65.13\% under the same retriever. This result shows that the retrieval pipeline remains effective without contextual selection, while the selector provides an additional gain. Selection still depends on the evidence available in the candidate pool and therefore complements rather than replaces expansion. Disabling expansion reduces Recall@5 by 10.95 points. Once expansion is enabled, the results from $L{=}2$ to $L{=}4$ remain within 0.35 points of $L{=}1$. This suggests that one expansion round captures most of the observed Recall@5 gain under the current setting. The result is consistent with latent hyperedges preserving the entities of an event within one retrieval unit, which reduces the amount of online expansion needed to connect evidence across chunks. Replacing hyperedge indexing with triple indexing causes a smaller decrease of 2.75 points. Triple-indexed SAG still exceeds HippoRAG~2 by 12.48 points, showing that SAG's advantage cannot be attributed to the hyperedge representation alone.

The remaining sweeps determine the default operating point. Increasing $K_{\mathrm{cand}}$ from 50 to 100 improves Recall@5 by 1.01 points, while increasing it from 100 to 500 adds only another 1.88 points despite a fivefold larger candidate budget. The runtime analysis in Appendix~\ref{appendix:efficiency} shows the corresponding cost. Moving from 50 to 100 candidates improves F1 by 0.74 points but increases per-query token consumption by 40.1\% and latency by 10.4\%. We therefore use $K_{\mathrm{cand}}=100$ as the accuracy-oriented configuration for the main experiments, while $K_{\mathrm{cand}}=50$ provides a more efficient alternative. The dual-path sweep also clarifies how the final output should be divided. Semantic retrieval alone reaches 56.23\% Recall@5 at $K_{\mathrm{event}}=0$, whereas the default setting with $K_{\mathrm{event}}=5$ reaches 80.36\%. Increasing the structural cap beyond five does not improve Recall@5. The best result is therefore obtained when structural selection retrieves evidence-chain events and semantic retrieval fills the remaining output slots.

\begin{figure}[htbp]
\centering
\begin{subfigure}[t]{0.315\linewidth}
\centering
\includegraphics[width=\linewidth]{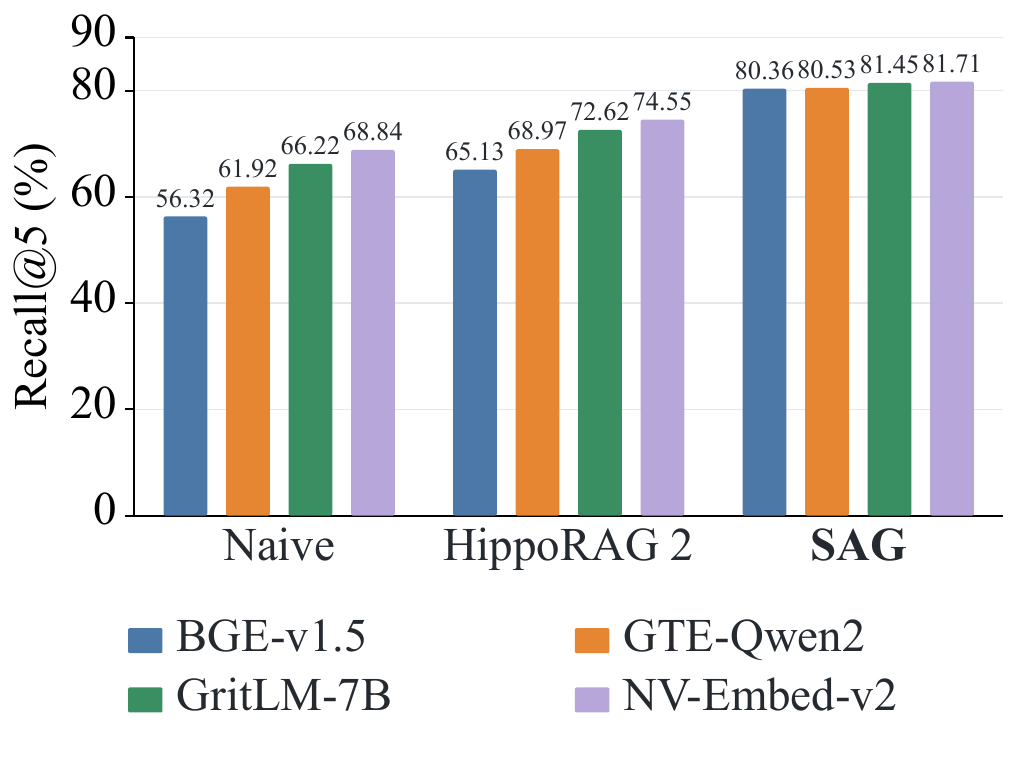}
\caption{Embedding Models}\label{fig:embedding-robustness}
\end{subfigure}\hfill
\begin{subfigure}[t]{0.315\linewidth}
\centering
\includegraphics[width=\linewidth]{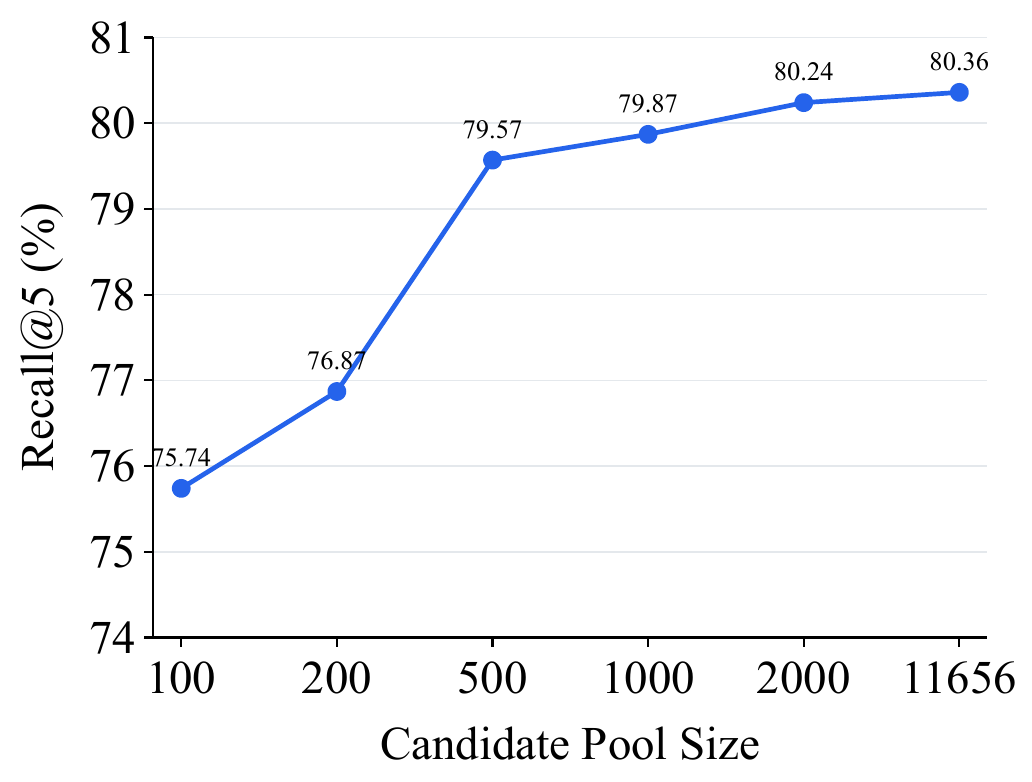}
\caption{Candidate Pool}\label{fig:candidate-pool-scaling}
\end{subfigure}\hfill
\begin{subfigure}[t]{0.315\linewidth}
\centering
\includegraphics[width=\linewidth]{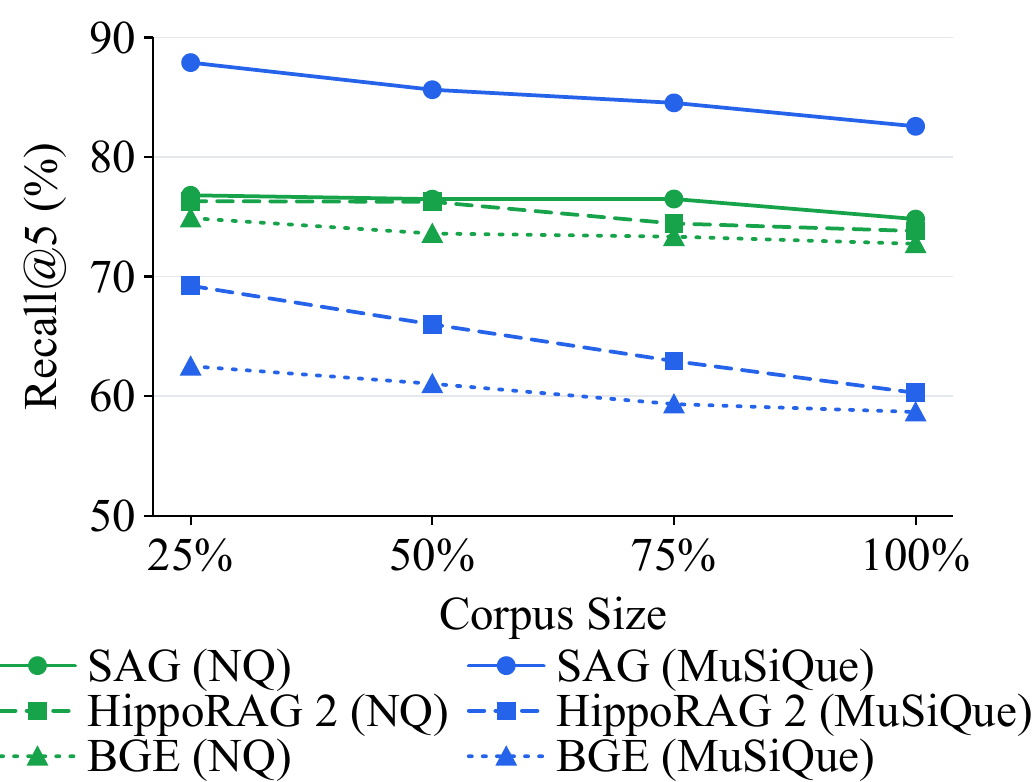}
\caption{Corpus Expansion}\label{fig:incremental}
\end{subfigure}
\caption{Robustness and scaling analyses. (a) Sensitivity of
four embedding models on MuSiQue. (b) Candidate-pool ablation on MuSiQue, from
$K_{\mathrm{pool}}=100$ to the full pool of 11,656 events. (c) Robustness of SAG, HippoRAG~2, and BGE to irrelevant corpus expansion on NQ and MuSiQue.}
\label{fig:robustness-scaling}
\end{figure}

\textbf{SAG remains robust across embedding models.}
We next examine whether SAG's retrieval advantage depends on the choice of embedding model (Figure~\ref{fig:embedding-robustness}). Replacing NV-Embed-v2 with BGE-v1.5 reduces HippoRAG~2 Recall@5 from 74.55\% to 65.13\%, a loss of 9.42 points, whereas SAG decreases from 81.71\% to 80.36\%, a loss of only 1.35 points. SAG with BGE-v1.5 still outperforms HippoRAG~2 with NV-Embed-v2. This difference is consistent with how the two systems use embeddings. HippoRAG~2 initializes personalized PageRank with embedding similarity, so changes in seed quality can affect downstream graph scores. SAG uses embeddings for seeding and coarse ranking, after which exact SQL joins expand the candidate set with connected events without re-applying embedding similarity at each step. The substantially smaller performance drop shows that SAG remains robust across embedding models, supporting the conclusion that its retrieval advantage arises primarily from the structured expansion mechanism rather than encoder strength.

\textbf{Scaling Up with Fixed Query-Time Budgets.}
At the current benchmark scale, SAG does not yet show a clear end-to-end cost or latency advantage (Appendix~\ref{appendix:efficiency}). Its scaling property is that only the inexpensive Elasticsearch HNSW search operates over the corpus-level index. HNSW retrieves the top $K_{\mathrm{pool}}$ chunks without exhaustive vector comparison, while entity expansion, SQL joins, and LLM selection remain within this query-specific working set. The costly LLM stage processes at most $K_{\mathrm{cand}}$ events. On MuSiQue, $K_{\mathrm{pool}}=500$ achieves 79.57\% Recall@5, compared with 80.36\% for all 11{,}656 chunks, retaining approximately 99\% of the full-pool result while activating less than 5\% of the corpus (Figure~\ref{fig:candidate-pool-scaling}). As the corpus grows by several orders of magnitude, $K_{\mathrm{pool}}$ may increase modestly to preserve recall, but the additional HNSW retrieval cost remains negligible relative to LLM inference. Keeping $K_{\mathrm{cand}}$ fixed therefore prevents the dominant query-time cost from growing proportionally with corpus size. SAG thus confines its expensive computation to a bounded, query-specific working set. We next examine whether retrieval quality remains stable as the corpus grows.

\textbf{SAG remains robust under continual corpus growth.}
We fix one quarter of NQ and MuSiQue for evaluation and incrementally add the remaining corpus, tracking Recall@5 as the corpus grows (Figure~\ref{fig:incremental}). On the single-hop NQ benchmark, SAG decreases only modestly from 76.80\% to 74.81\% as the corpus grows from 25\% to 100\%, and remains the strongest method at every corpus size. This robustness also holds on the more challenging multi-hop MuSiQue benchmark. SAG decreases from 87.90\% to 82.57\%, a loss of 5.33 points, whereas HippoRAG~2 falls from 69.23\% to 60.27\%, a loss of 8.96 points. Corpus growth introduces entity-overlapping distractors that can displace bridge passages, making complete evidence chains more difficult to recover. SAG's slower degradation is consistent with its append-only event--entity index and query-local SQL joins, which limit the influence of unrelated additions on existing retrieval paths. These results show that SAG remains stable on the simpler single-hop task while preserving a clear advantage on the more demanding multi-hop task.
\subsection{Limitations}
\label{sec:limitations}

SAG has two current limitations. First, it normalizes and deduplicates entity strings but does not resolve aliases. It therefore cannot recognize that ``Apple Inc.'' and ``Apple'' refer to the same entity, and connections that depend on matching these surface forms may be lost. A lightweight alias table could link common full names, abbreviations, and aliases without requiring full entity disambiguation or compromising incremental writes.

The second limitation concerns temporal updates. SAG's index is append-only by design, which supports continual ingestion but provides no mechanism for revising or retiring stale events. This is sufficient for a static retrieval corpus, but an agent memory must also represent facts, preferences, and task states that change over time. Versioned events, informed by bitemporal knowledge graphs~\citep{rasmussen2025zep,rasmussen2025atom}, are a natural next step toward supporting knowledge that can be updated while preserving its history.

\section{Conclusion}

We have proposed SAG, a structured retrieval architecture that represents chunks as semantically complete events and dynamically activates query-specific latent hyperedges through shared entities. This design turns indexed data into structured context for multi-hop retrieval. Across three multi-hop benchmarks, SAG achieves the strongest average retrieval and QA performance under unified evaluation settings, with its largest gains on MuSiQue. The ablation results show that event representation, query-time expansion, and LLM-based final selection each contribute to retrieval quality. The robustness experiments further show that SAG varies less across embedding models and degrades more slowly as the corpus grows. Fixed activation budgets bound the downstream SQL expansion and LLM selection workloads, while append-only indexing supports continual writes without global graph reconstruction. Future work will examine larger-scale deployments and extend SAG with alias resolution and versioned updates toward a memory substrate for agents.

\subsection*{AI use statement}

Large language models were used during the preparation of this manuscript for writing assistance, including grammar checking, sentence-level rephrasing, and LaTeX formatting. All scientific content, experimental design, data analysis, and technical claims were produced entirely by the authors. The authors take full responsibility for the correctness and integrity of all reported results.

\subsection*{Ethics statement}

The experiments use public multi-hop QA datasets and collect no new human-subject
data. Because SAG uses language models to extract and link entities, errors or
biases may produce misleading associations, while deployment on private corpora
may expose sensitive cross-document relationships. Such deployments should
enforce data provenance, access control, and human review in high-impact settings.

\subsection*{Reproducibility statement}

All experiments and results reported in the main paper and appendices can be
reproduced using the benchmark code, evaluation scripts, and configurations
available at \url{https://github.com/Zleap-AI/SAG-Benchmark}. A separate
engineering implementation of SAG is available at
\url{https://github.com/Zleap-AI/SAG}. The paper and appendices specify the
datasets, models, evaluation settings, ablations, costs, and prompts. Exact
values involving hosted Qwen models may vary with provider updates and
nondeterministic inference.

\bibliography{iclr2027_conference}
\bibliographystyle{iclr2027_conference}

\clearpage
\appendix
\raggedbottom
\section{Latent Hyperedge Formalization}
\label{appendix:formalization}

This appendix states in full the formalization summarized in Section~\ref{subsec:hyperedge}.

\textbf{Definition 1 (Latent Hyperedge).} Let $V$ be the set of entity vertices extracted from a corpus and $H$ the set of indexed events (hyperedges), each with a stable identifier. Every event $h\in H$ is represented as a labeled hyperedge incident to an entity set $V(h)=\{v_1,\ldots,v_n\}\subseteq V$, where $n\geq 1$. Because one latent hyperedge simultaneously connects $n$ entities, it preserves the $n$-ary co-occurrence structure of one chunk as an atomic retrieval unit. A single event ``Company A acquired Company B for \$X on Date D'' links acquirer, target, price, and date in one unit. A direct binary decomposition requires multiple records and retains their common acquisition context only if it introduces an auxiliary event node or an equivalent reification mechanism. In SAG, the event description labels the hyperedge, while a stable event-to-chunk mapping $\phi: H \to D$ preserves the original chunk as the final evidence unit. This definition concerns the representation of each event; it does not define SAG as a materialized hypergraph.

\textbf{Proposition 1 (Lossless Incidence Encoding).} Define the incidence set
\begin{equation}
I=\{(h,v)\in H\times V\mid v\in V(h)\}.
\label{eq:bipartite}
\end{equation}
Assuming that every incidence row is retained and event identifiers are unique, the event-entity SQL table of Section~\ref{subsec:index} stores $I$ exactly and is therefore a lossless bipartite representation of the latent-hyperedge incidence relation \citep{feng2025hypergraphrag}. For every event,
\begin{equation}
V(h)=\{v\in V\mid(h,v)\in I\}.
\label{eq:recover-hyperedge}
\end{equation}
Thus, grouping table rows by $h$ reconstructs each hyperedge, while enumerating the pairs in all reconstructed hyperedges recovers the table. This proposition concerns storage after extraction; it makes no claim that mapping a source chunk to an event and entities preserves all source information.

\textbf{Proposition 2 (Query-Time Dynamic Activation).} Fix an index and retrieval configuration, and let $S_0=R$ be the query-dependent seed event set used in Section~\ref{subsec:expansion}. For unpruned expansion, define
\begin{equation}
F_{t+1}=\bigcup_{h\in S_t}V(h),
\qquad
S_{t+1}=\{h\in H\mid V(h)\cap F_{t+1}\neq\varnothing\}.
\label{eq:local}
\end{equation}
One expansion round computes $S_t\to F_{t+1}\to S_{t+1}$ through two SQL joins, corresponding to two traversals in the bipartite incidence graph $B=(V\cup H,I)$. Consequently, after $L$ rounds the unpruned candidate events are exactly the event vertices within bipartite distance at most $2L$ from $S_0$. Frontier pruning only removes entities or events from these reachable sets, so practical SAG activates a budget-constrained, query-scoped set of incident latent hyperedges rather than the complete neighborhood. The activation is dynamic because this set is assembled for the query from its seed events under the fixed retrieval configuration, not because new corpus-level hyperedges are created at query time.

\section{Corpus-Level Connectivity Analysis}
\label{appendix:connectivity}

The formalization above describes one event in isolation. We next examine how latent hyperedges organize an entire corpus. A three-layer view makes the indexing structure explicit: each source chunk has a one-to-one mapping to a semantically complete event, while the event has a one-to-many mapping to its indexing entities. Entities that recur across events create paths between otherwise separate chunks. Thus, SAG does not connect chunks by replacing their text with graph fragments; it connects them through an intermediate event layer while retaining every original chunk as the evidence carrier.

\paragraph{Connection capacity of an event.}
Let $n_h=|V(h)|$ be the number of entities attached to event $h$. If the hyperedge is projected onto an entity--entity graph, its members admit
\begin{equation}
P(h)=\binom{n_h}{2}
\label{eq:pair-capacity}
\end{equation}
distinct pairwise co-memberships. We call $P(h)$ the \emph{pairwise connection capacity} of the event. It is a representational quantity rather than a count of physically stored edges: SAG stores only the $n_h$ incidence rows $(h,v)$ and recovers the co-memberships by grouping on the event identifier. Across a corpus, the corresponding capacity is
\begin{equation}
P_{\mathrm{corp}}=\sum_{h\in H}\binom{n_h}{2}.
\label{eq:corpus-pair-capacity}
\end{equation}
A star-shaped binary decomposition needs $n_h-1$ records to connect the same $n_h$ entities to a designated anchor, giving the corpus-level amplification factor
\begin{equation}
A_{\mathrm{corp}}=
\frac{\displaystyle\sum_{h\in H}\binom{n_h}{2}}
     {\displaystyle\sum_{h\in H,\,n_h\geq 2}(n_h-1)}.
\label{eq:connection-amplification}
\end{equation}
This factor should not be read as a universal storage advantage over every possible triple schema. It quantifies how many pairwise co-memberships are jointly available from one labeled event unit relative to a fixed anchor-based binary decomposition. An arbitrary triple representation can preserve the same information only by choosing and maintaining an explicit reification scheme; independently indexed triples do not, by themselves, retain the common event boundary.

\paragraph{Cross-framework comparison on MuSiQue.}
Following the graph-indexing evaluation protocol of \citet{xiang2025use}, the evaluation converts each framework's native index into an undirected graph and applies the same component and degree calculations. The resulting graphs are not isomorphic representations: SAG uses event--entity incidences, GraphRAG and HyperGraphRAG use entity--entity links, LightRAG uses entity--relation incidences, and HippoRAG~2 uses passage--entity links. Raw node and link counts must therefore be read as native index scale rather than directly interchangeable units. Component count, giant-component coverage, and isolation rate nevertheless reveal how fragmented each native retrieval structure is. Here and throughout the paper, GraphRAG refers to the official Microsoft GraphRAG implementation (release v2.0.0).

\paragraph{Metric definitions and interpretation.}
For a framework's undirected native index $G=(V_G,E_G)$:

\noindent\textbf{Nodes.} The total number of vertices $|V_G|$ in the native index. A vertex may represent an entity, event, relation, or passage depending on the framework.

\noindent\textbf{Links.} The total number of edges or incidence links $|E_G|$. Link semantics likewise depend on the native index schema.

\noindent\textbf{Avg.\ degree.} The mean number of incident links per vertex, $\bar d=2|E_G|/|V_G|$. It measures connection intensity and traversal burden, not retrieval accuracy.

\noindent\textbf{Components.} The number of connected subgraphs that are mutually unreachable. Fewer components indicate less fragmentation, but the value should be interpreted together with graph size.

\noindent\textbf{Giant comp.} The number of vertices $|C_{\max}|$ in the largest connected component and its percentage of all nodes, $100\times |C_{\max}|/|V_G|$. A higher percentage indicates that more of the index belongs to one potentially reachable region.

\noindent\textbf{Isolated.} The number of degree-zero vertices and their percentage of all nodes, $100\times |\{v\in V_G\mid\deg(v)=0\}|/|V_G|$. These vertices cannot be reached from any other vertex through graph traversal.

Nodes and links describe native index scale rather than semantic richness because their meanings differ across frameworks. For the connectivity metrics, lower component and isolation values indicate less disconnected content, while higher giant-component coverage indicates broader potential reachability.

We omit density, clustering coefficient, and whole-graph diameter from the primary comparison. Density is dominated by graph scale, triangle-based clustering is structurally zero for bipartite indices such as SAG, and whole-graph diameter is infinite whenever an index is disconnected. These descriptors therefore do not directly answer whether chunks participate in reachable cross-document structures.

\begin{table}[H]
\caption{Native index scale and connectivity on MuSiQue. ``Index relation'' identifies the vertex and link semantics used by each framework. Since these semantics differ, nodes and links are descriptive scale statistics; components, giant-component coverage, and isolation quantify fragmentation within each native index. Percentages in parentheses use the Nodes column as the denominator.}
\label{table-framework-connectivity}
\centering
\small
\resizebox{\linewidth}{!}{%
\begin{tabular}{llrrrrrr}
\toprule
Method & Index relation & Nodes & Links & Avg.\ degree & Components & Giant comp. & Isolated \\
\midrule
GraphRAG & Entity--entity & 128,189 & 106,489 & 1.66 & 66,472 & 58,815 (45.9\%) & 65,506 (51.1\%) \\
LightRAG & Entity--relation & 82,321 & 116,555 & 2.83 & 9,277 & 70,196 (85.3\%) & 8,512 (10.3\%) \\
HyperGraphRAG & Entity--entity & 186,897 & 272,524 & 2.92 & 4,600 & 171,196 (91.6\%) & 485 (0.3\%) \\
HippoRAG~2 & Passage--entity & 97,043 & 570,458 & 11.76 & 96 & 96,773 (99.7\%) & 59 (0.1\%) \\
SAG & Event--entity & 95,341 & 126,540 & 2.65 & 400 & 92,206 (96.7\%) & 1 (0.001\%) \\
\bottomrule
\end{tabular}%
}
\end{table}

GraphRAG has 166 times as many connected components as SAG, LightRAG has 23.2 times as many, and HyperGraphRAG has 11.5 times as many. SAG's largest component covers 96.7\% of the native index, compared with 45.9\%, 85.3\%, and 91.6\%, respectively, while its isolated-vertex rate is effectively zero. These differences show that repeated entities connect semantically complete events into a broad cross-chunk substrate rather than leaving them as local records.

HippoRAG~2 occupies the opposite end of the connectivity spectrum: it has only 96 components and 99.7\% giant-component coverage, but uses 570,458 links, 4.5 times SAG's count, and raises the average degree from 2.65 to 11.76. Yet its MuSiQue Recall@5 is 65.13\%, compared with SAG's 80.36\%. The comparison therefore does not support the claim that more native links are inherently better. It supports a narrower claim: SAG attains low fragmentation and high reachability at moderate degree, while its query-time event--entity traversal turns that structure into more useful retrieved evidence.

\begin{table}[H]
\caption{Connectivity of SAG's event--entity incidence index on the three evaluation corpora. Vertices include events and entities; incidence rows are event--entity links. Lower component and isolated-vertex counts indicate less fragmentation, while larger giant-component coverage indicates greater corpus-level reachability. Percentages in parentheses use the Vertices column as the denominator.}
\label{table-sag-connectivity}
\centering
\small
\setlength{\tabcolsep}{5pt}
\resizebox{\linewidth}{!}{%
\begin{tabular}{ccccccc}
\toprule
Dataset & Vertices & Incidences & Avg.\ degree & Components & Giant comp. & Isolated vertices \\
\midrule
MuSiQue          & 95,341  & 126,540 & 2.65 & 400   & 92,206 (96.7\%) & 1 (0.001\%) \\
HotpotQA         & 100,570 & 186,634 & 3.71 & 5,869 & 93,593 (93.1\%) & 5,787 (5.8\%) \\
2WikiMultiHopQA  & 57,812  & 75,399  & 2.61 & 272   & 56,270 (97.3\%) & 0 (0.0\%) \\
\bottomrule
\end{tabular}
}
\end{table}

\paragraph{SAG connectivity across datasets.}
The largest connected component covers more than 93\% of the index on every dataset and more than 96\% on MuSiQue and 2WikiMultiHopQA. This does not mean that every pair of chunks is relevant to the same query, nor that traversal should be global. It shows that shared entities provide a broad substrate of possible cross-chunk routes from which SAG activates a small query-conditioned neighborhood. The moderate average degree of 2.61--3.71 further distinguishes reachability from indiscriminate density: the index is broadly connected without requiring every event to become a high-degree hub.

\paragraph{Connection structure and retrieval.}
The matched indexing ablation in Table~\ref{table-ablation} holds the corpus, retrieval pipeline, and downstream model fixed and changes only the representation. Replacing independently indexed triples with latent hyperedges raises Recall@2 from 61.83\% to 63.62\%, Recall@5 from 77.61\% to 80.36\%, and Recall@10 from 81.54\% to 83.37\% on MuSiQue. The gain grows beyond the first rank, which is consistent with the proposed mechanism: the event layer contributes primarily by exposing additional evidence through shared-entity paths rather than by improving the highest-scoring direct match. Disabling expansion produces a larger Recall@5 drop, from 80.36\% to 69.41\%, confirming that the available incidence structure matters only when the online procedure traverses it. Together, the corpus statistics and matched ablations separate three claims: latent hyperedges provide higher-order connection capacity, repeated entities turn that capacity into cross-chunk reachability, and query-time expansion converts the reachable structure into retrieved evidence.

\section{Qualitative Examples of Query-Time Hyperedge Activation}
\label{appendix:trace}

The intermediate states of the retrieval procedure in Section~\ref{subsec:expansion} form an auditable execution chain:
\begin{equation}
q \rightarrow Q \rightarrow \widehat{Q} \rightarrow R \rightarrow C
\rightarrow \widehat{C} \rightarrow D_{\mathrm{out}}.
\label{eq:retrieval-trace}
\end{equation}

\begin{figure}[H]
\centering
\begin{minipage}[t]{0.58\linewidth}
\vspace{0pt}
\scriptsize
\renewcommand{\arraystretch}{1.18}
\begin{tabular}{@{}>{\raggedright\arraybackslash\bfseries}p{0.24\linewidth}
                @{\hspace{0.8em}}>{\raggedright\arraybackslash}p{0.70\linewidth}@{}}
\toprule
Query &
How many times did the plague occur in the city where the painter of \textit{The Bacchanal of the Andrians} died? \\
\midrule
Gold Docs &
1.~The Bacchanal of the Andrians;
2.~Black Death;
3.~The Martyrdom of Saint Lawrence (Titian) \\
\midrule
Extracted entity &
The Bacchanal of the Andrians \\
\midrule
Expanded entity &
Venice \\
\midrule
Expanded events &
1.~Venice;
\textbf{2.~The Martyrdom of Saint Lawrence (Titian)};
3.~Venice \\
\midrule
Recalled events (Top-5) &
\textbf{1.~Black Death};
\textbf{2.~The Bacchanal of the Andrians};
3.~Venice;
\textbf{4.~The Martyrdom of Saint Lawrence (Titian)};
5.~Venice \\
\bottomrule
\end{tabular}
\end{minipage}
\hfill
\begin{minipage}[t]{0.39\linewidth}
\vspace{0pt}
\centering
\includegraphics[width=\linewidth]{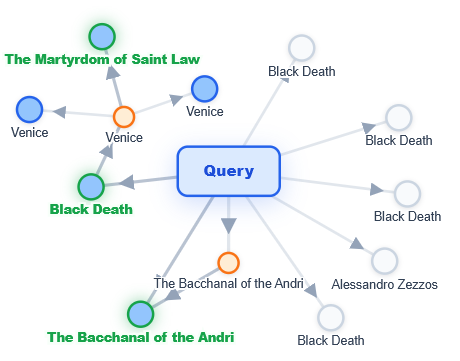}
\par\vspace{0.1em}
\includegraphics[width=\linewidth]{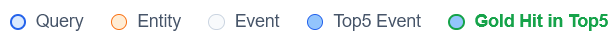}
\end{minipage}
\caption{Example with a good retrieval result. The initial paths recover two gold events, and the shared entity \textit{Venice} activates an additional latent hyperedge that recovers the third gold document. Bold titles mark gold documents in the corresponding table rows.}
\label{fig:good-query-trace}
\end{figure}

\begin{figure}[H]
\centering
\begin{minipage}[t]{0.58\linewidth}
\vspace{0pt}
\scriptsize
\renewcommand{\arraystretch}{1.18}
\begin{tabular}{@{}>{\raggedright\arraybackslash\bfseries}p{0.24\linewidth}
                @{\hspace{0.8em}}>{\raggedright\arraybackslash}p{0.70\linewidth}@{}}
\toprule
Query &
When did the city where Greenwood Laboratory School is located become capital of the state where the screenwriter of \textit{The Poor Boob} was born? \\
\midrule
Gold Docs &
1.~Greenwood Laboratory School;
2.~Margaret Mayo (playwright);
3.~The Poor Boob;
4.~Springfield, Illinois \\
\midrule
Extracted entities &
The Poor Boob;
Greenwood Laboratory School \\
\midrule
Expanded entity &
None \\
\midrule
Expanded events &
None \\
\midrule
Recalled events (Top-5) &
\textbf{1.~Greenwood Laboratory School};
2.~Maine;
\textbf{3.~The Poor Boob};
4.~Washington (state);
5.~Oklahoma \\
\bottomrule
\end{tabular}
\end{minipage}
\hfill
\begin{minipage}[t]{0.39\linewidth}
\vspace{0pt}
\centering
\includegraphics[width=\linewidth]{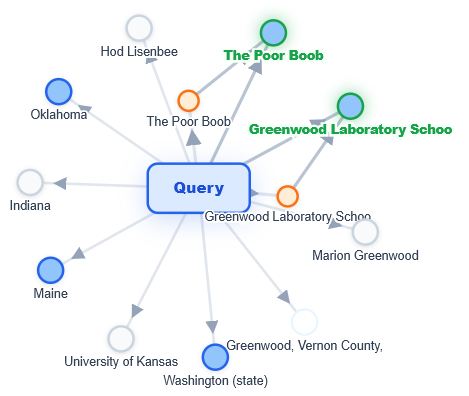}
\end{minipage}
\caption{Example with a poor retrieval result. The initial paths recover two gold events, but no shared entity is available to activate latent hyperedges leading to the remaining evidence. Bold titles denote gold documents in the Top-5 result.}
\label{fig:bad-query-trace}
\end{figure}

Figures~\ref{fig:good-query-trace} and~\ref{fig:bad-query-trace} pair the recorded intermediate outputs with their retrieval visualizations. The left-hand tables expose the extracted query entities, seed events returned by the lexical and vector paths, shared entities available for joining, and latent hyperedges activated during expansion. The right-hand diagrams visualize the corresponding query-centered neighborhoods. Both examples retrieve relevant evidence, but they differ in how effectively query-time dynamic hyperedge activation extends the initial result. The legend directly below the good-example diagram applies to both diagrams.

The good example directly illustrates query-time dynamic hyperedge activation. The initial lexical and vector paths recover \textit{Black Death} and \textit{The Bacchanal of the Andrians}. SAG then discovers the shared entity \textit{Venice} through SQL joins and activates its incident latent hyperedges, bringing \textit{The Martyrdom of Saint Lawrence (Titian)} into the final Top-5 result. The poor example exposes the complementary boundary. Although direct retrieval finds \textit{Greenwood Laboratory School} and \textit{The Poor Boob}, the returned events provide no retained shared entity that can activate a route toward \textit{Margaret Mayo (playwright)} and \textit{Springfield, Illinois}. The contrast isolates the role of dynamic hyperedges: they add useful cross-document evidence when the seed events reveal a discriminative bridge, while the absence of such a bridge leaves the result dependent on direct retrieval.

\section{Full Retrieval Results}
\label{appendix:full-results}

\begin{table}[H]
\caption{Full retrieval results (Recall@1/2/5/10) on three multi-hop benchmarks. All structure-augmented methods use BGE-Large-EN-v1.5 as the retriever; Qwen3.6-Flash is used as the reader.}
\label{table-full-results}
\centering
\resizebox{\linewidth}{!}{%
\begin{tabular}{lcccccccccccc}
\toprule
& \multicolumn{4}{c}{MuSiQue} & \multicolumn{4}{c}{2WikiMultiHopQA} & \multicolumn{4}{c}{HotpotQA} \\
\cmidrule(lr){2-5}\cmidrule(lr){6-9}\cmidrule(lr){10-13}
& R@1 & R@2 & R@5 & R@10 & R@1 & R@2 & R@5 & R@10 & R@1 & R@2 & R@5 & R@10 \\
\midrule
\multicolumn{13}{c}{\cellcolor{grouphdr}\textit{Simple Baselines}} \\
BM25 & 24.59 & 32.96 & 42.11 & 49.69 & 36.85 & 54.30 & 63.78 & 67.90 & 38.75 & 56.45 & 73.05 & 85.40 \\
Contriever & 24.50 & 33.54 & 45.32 & 53.58 & 34.40 & 48.60 & 59.25 & 65.20 & 39.30 & 57.45 & 74.85 & 82.75 \\
\multicolumn{13}{c}{\cellcolor{grouphdr}\textit{Large Embedding Models}} \\
BGE-Large-EN-v1.5 & 29.42 & 41.68 & 56.32 & 64.08 & 41.48 & 62.28 & 69.83 & 73.75 & 45.45 & 76.20 & 89.20 & 93.35 \\
GTE-Qwen2-7B & \underline{35.19} & 48.06 & 61.92 & 70.80 & \underline{43.45} & 65.58 & 74.18 & 78.63 & 46.40 & 73.95 & 88.45 & 93.55 \\
GritLM-7B & 34.86 & 50.70 & 66.22 & 75.19 & 41.85 & 66.23 & 75.62 & 79.40 & 46.40 & 81.15 & 93.40 & \underline{97.20} \\
NV-Embed-v2 & 34.02 & \underline{51.08} & \underline{68.84} & \underline{77.27} & 43.23 & 68.93 & 77.00 & 80.75 & \underline{46.90} & \underline{85.55} & \underline{95.25} & \textbf{97.70} \\
\multicolumn{13}{c}{\cellcolor{grouphdr}\textit{Structure-Augmented RAG}} \\
HippoRAG 2 & 30.65 & 49.52 & 65.13 & 73.76 & 42.38 & \underline{76.55} & \underline{90.35} & \underline{93.40} & 44.40 & 78.35 & 94.35 & 97.15 \\
\rowcolor{tableoneblue} SAG (Ours) & \textbf{36.82} & \textbf{63.62} & \textbf{80.36} & \textbf{83.37} & \textbf{43.81} & \textbf{83.93} & \textbf{93.34} & \textbf{93.57} & \textbf{47.80} & \textbf{91.55} & \textbf{96.50} & \textbf{97.70} \\
\bottomrule
\end{tabular}%
}
\end{table}

\section{Full QA Results}
\label{appendix:full-qa}

\begin{table}[H]
\caption{Full QA results (EM/F1) on three multi-hop benchmarks. All methods use Qwen3.6-Flash as the reader, and all structure-augmented methods use BGE-Large-EN-v1.5 as the retriever.}
\label{table-full-qa}
\centering
\resizebox{\linewidth}{!}{%
\begin{tabular}{lcccccccc}
\toprule
& \multicolumn{2}{c}{MuSiQue} & \multicolumn{2}{c}{2Wiki} & \multicolumn{2}{c}{HotpotQA} & \multicolumn{2}{c}{Avg} \\
\cmidrule(lr){2-3}\cmidrule(lr){4-5}\cmidrule(lr){6-7}\cmidrule(lr){8-9}
& EM & F1 & EM & F1 & EM & F1 & EM & F1 \\
\midrule
\multicolumn{9}{c}{\cellcolor{grouphdr}\textit{Simple Baselines}} \\
BM25~\citep{robertson1994bm25} & 26.10 & 35.19 & 52.40 & 56.26 & 52.90 & 64.64 & 43.80 & 52.03 \\
Contriever~\citep{izacard2022contriever} & 25.70 & 34.39 & 45.20 & 50.62 & 52.80 & 64.06 & 41.23 & 49.69 \\
\multicolumn{9}{c}{\cellcolor{grouphdr}\textit{Large Embedding Models}} \\
BGE-Large-EN-v1.5~\citep{xiao2023bge} & 33.00 & 42.80 & 56.10 & 61.49 & 62.40 & 74.81 & 50.50 & 59.70 \\
GTE-Qwen2-7B~\citep{li2023gte} & 34.70 & 44.60 & 59.90 & 64.91 & 61.10 & 73.04 & 51.90 & 60.85 \\
GritLM-7B~\citep{muennighoff2025gritlm} & 38.60 & 49.30 & 60.70 & 66.12 & 61.00 & 73.22 & 53.43 & 62.88 \\
NV-Embed-v2~\citep{lee2024nvembed} & 41.70 & 52.88 & 62.50 & 68.45 & \underline{65.40} & \underline{78.00} & 56.53 & 66.44 \\
\multicolumn{9}{c}{\cellcolor{grouphdr}\textit{Structure-Augmented RAG}} \\
GraphRAG~\citep{edge2024graphrag} & \underline{43.20} & \underline{54.14} & 64.70 & 72.08 & 63.70 & 76.29 & 57.20 & 67.50 \\
LightRAG~\citep{guo2025lightrag} & 33.60 & 46.66 & 59.00 & 65.75 & 61.20 & 73.09 & 51.27 & 61.83 \\
HyperGraphRAG~\citep{feng2025hypergraphrag} & 39.50 & 49.20 & 57.80 & 63.44 & 62.60 & 74.70 & 53.30 & 62.45 \\
HyperRAG~\citep{feng2026hyperrag} & 40.90 & 51.80 & \underline{69.40} & \underline{76.89} & 64.40 & 77.19 & \underline{58.23} & \underline{68.63} \\
HippoRAG~2~\citep{gutierrez2025hipporag2} & 37.90 & 47.49 & 67.00 & 74.34 & 65.20 & 77.24 & 56.70 & 66.36 \\
\rowcolor{tableoneblue} SAG (Ours) & \textbf{50.20} & \textbf{61.15} & \textbf{70.50} & \textbf{78.06} & \textbf{67.90} & \textbf{79.66} & \textbf{62.87} & \textbf{72.96} \\
\bottomrule
\end{tabular}%
}
\end{table}

\begingroup
\raggedbottom
\section{Full Ablation and Hyperparameter Results}
\label{appendix:ablation}

\setlength{\intextsep}{8pt}
\begin{table}[H]
\centering
\small
\caption{Ablation of SAG's key design choices on MuSiQue. The first row reports the complete default configuration. Each subsequent row changes the indicated stage while holding all remaining settings fixed. Bold marks the best result.}
\label{table-ablation}
\setlength{\tabcolsep}{4pt}
\begin{tabular}{llcccc}
\toprule
& & \multicolumn{4}{c}{MuSiQue} \\
\cmidrule(lr){3-6}
Stage & Configuration & R@1 & R@2 & R@5 & R@10 \\
\midrule
Default & SAG (Ours)\footnotemark & \textbf{36.82} & \textbf{63.62} & \textbf{80.36} & \textbf{83.37} \\
\midrule
Indexing & w/ Triple indexing & 35.66 & 61.83 & 77.61 & 81.54 \\
Expansion & w/o Expansion ($L=0$) & 35.70 & 57.75 & 69.41 & 74.76 \\
Final selection & w/ Qwen3-Reranker-8B & 32.12 & 48.97 & 67.11 & 76.51 \\
\bottomrule
\end{tabular}
\end{table}
\footnotetext{Defaults for the ablated stages: hyperedge indexing, one-hop expansion ($L=1$), and Qwen3.6-Flash final selection.}

\begin{table}[H]
\centering
\small
\caption{Hyperparameter sweeps on MuSiQue. Expansion depth varies $L$; $L=0$ corresponds to the w/o Expansion ablation in Table~\ref{table-ablation}, while $L=1$ is the default SAG setting. Candidate budget varies $K_{\mathrm{cand}}$. In the dual-path sweep, $K_{\mathrm{event}}$ is the maximum number of events selected by the LLM, and the semantic path fills the remaining slots up to $K_{\mathrm{out}}=10$. All other settings remain at their defaults. Bold marks the best value within each group.}
\label{table-hyperparameters}
\setlength{\tabcolsep}{4pt}
\begin{tabular}{llcccc}
\toprule
& & \multicolumn{4}{c}{MuSiQue} \\
\cmidrule(lr){3-6}
Parameter & Configuration & R@1 & R@2 & R@5 & R@10 \\
\midrule
\multirow{5}{*}{Expansion depth} & $L=0$ & 35.70 & 57.75 & 69.41 & 74.76 \\
 & $L=1$ (default) & \textbf{36.82} & 63.62 & 80.36 & 83.37 \\
 & $L=2$ & 36.17 & \textbf{64.05} & 80.47 & 84.00 \\
 & $L=3$ & 36.03 & 63.42 & 80.40 & 83.79 \\
 & $L=4$ & 36.27 & 63.83 & \textbf{80.71} & \textbf{84.14} \\
\midrule
\multirow{4}{*}{Candidate budget} & $K_{\mathrm{cand}}=50$ & 36.53 & 62.80 & 79.35 & 81.69 \\
 & $K_{\mathrm{cand}}=100$ (default) & 36.82 & 63.62 & 80.36 & 83.37 \\
 & $K_{\mathrm{cand}}=200$ & \textbf{36.99} & \textbf{65.12} & 81.06 & 84.47 \\
 & $K_{\mathrm{cand}}=500$ & 36.46 & 64.72 & \textbf{82.24} & \textbf{86.56} \\
\midrule
\multirow{7}{*}{Dual-path output} & $K_{\mathrm{event}}=0$ (semantic only) & 29.37 & 41.56 & 56.23 & 63.77 \\
 & $K_{\mathrm{event}}=2$ & 36.62 & \textbf{64.56} & 73.52 & 77.64 \\
 & $K_{\mathrm{event}}=4$ & 36.61 & 63.73 & 79.80 & 82.63 \\
 & $K_{\mathrm{event}}=5$ (default) & \textbf{36.82} & 63.62 & \textbf{80.36} & 83.37 \\
 & $K_{\mathrm{event}}=6$ & 36.71 & 63.83 & 79.93 & 83.40 \\
 & $K_{\mathrm{event}}=8$ & 36.00 & 62.86 & 80.14 & \textbf{83.57} \\
 & $K_{\mathrm{event}}=10$ & 36.46 & 63.48 & 79.80 & 83.18 \\
\bottomrule
\end{tabular}
\end{table}
\endgroup

\section{Runtime, Token Consumption, and Scaling}
\label{appendix:efficiency}

\begin{table}[H]
\centering
\small
\caption{Runtime, token consumption, and quality on MuSiQue. Offline totals cover
11,656 corpus chunks; online values are normalized over 1,000 queries. End-to-end
time is retrieval time plus question-answering time. F1 scores for non-SAG methods
are from Table~\ref{table-full-qa}. For SAG, $K_{\mathrm{cand}}$ variants illustrate
the cost--quality trade-off; the offline indexing cost (4.57\,h, 28.26\,M tokens) is
shared across all variants. The final Tokens column reports total input and output
tokens per query across retrieval and QA.}
\label{table-efficiency}
\setlength{\tabcolsep}{2.5pt}
\begin{tabular}{ll*{7}{c}}
\toprule
& & & \multicolumn{2}{c}{Offline indexing} & \multicolumn{4}{c}{Online querying (per query)} \\
\cmidrule(lr){4-5}\cmidrule(lr){6-9}
Method & & F1 & Time (h) & Tokens (M) & Retrieval (s) & QA (s) & End-to-end (s) & Tokens \\
\midrule
RAG (BGE)     && 42.80 & 0.00  & 0      & 0.10  & 1.75  & 1.85  & 1,727  \\
GraphRAG      && 54.14 & 10.15 & 125.12 & 0.23  & 12.56 & 12.79 & 11,438 \\
LightRAG      && 46.66 & 25.94 & 103.14 & 11.97 & 18.13 & 30.10 & 9,974  \\
HyperGraphRAG && 49.20 & 16.21 & 216.11 & 4.45  & 23.54 & 27.99 & 32,536 \\
HippoRAG~2    && 47.49 & 2.30  & 15.93  & 0.66  & 8.50  & 9.16  & 4,788  \\
\midrule
\multirow{2}{*}{SAG (Ours)}
& $K_{\mathrm{cand}}\!=\!50$  & 60.41 & \multirow{2}{*}{4.57} & \multirow{2}{*}{28.26} & 15.93 & 6.75 & 22.68 & 8,236  \\
& $K_{\mathrm{cand}}\!=\!100$ & \textbf{61.15} & & & 18.67 & 6.36 & 25.03 & 11,536 \\
\bottomrule
\end{tabular}
\vspace{2pt}
\end{table}

\textbf{Offline construction cost.} SAG's index requires 4.57\,h and
28.26\,M tokens to build. Relative to GraphRAG, LightRAG, and HyperGraphRAG,
this reduces indexing time by 55.0--82.4\% and indexing tokens by
72.6--86.9\%. HippoRAG~2 remains the cheaper offline baseline at 2.30\,h and
15.93\,M tokens; SAG therefore reduces, but does not eliminate, the offline
cost of structure augmentation.

\textbf{Online cost--quality frontier.} At $K_{\mathrm{cand}}=50$, SAG is
simultaneously more accurate, faster end to end, and less token-intensive than
LightRAG and HyperGraphRAG: it obtains 60.41 F1 in 22.68\,s with 8,236 tokens,
versus 46.66/30.10\,s/9,974 tokens and 49.20/27.99\,s/32,536 tokens,
respectively. Against GraphRAG and HippoRAG~2, the comparison is instead a
quality--latency trade-off: SAG gains 6.27 and 12.92 F1 points, respectively,
but takes longer end to end. The stage breakdown explains where this time is
spent. SAG's 15.93\,s retrieval-and-selection stage is the bottleneck, whereas
its 6.75\,s QA stage is faster than those of all four structure-augmented
baselines (8.50--23.54\,s). Thus, SAG moves computation from answer generation
to evidence construction rather than being the fastest system overall.

\textbf{Return on LLM computation.} The controlled substitutions show what
the online selection cost buys. In Table~\ref{table-ablation}, replacing
contextual Qwen3.6-Flash selection with the pointwise Qwen3-Reranker-8B lowers
Recall@5 from 80.36\% to 67.11\%, a 13.25-point drop. In
Table~\ref{table-hyperparameters}, setting $K_{\mathrm{event}}=0$ and thereby
removing LLM-selected structural output lowers Recall@5 to 56.23\%, 24.13
points below the default dual-path configuration. Within SAG, increasing
$K_{\mathrm{cand}}$ from 50 to 100 adds only 0.74 F1 points while increasing
per-query token consumption by 40.1\% and end-to-end latency by 10.4\%.
Consequently, $K_{\mathrm{cand}}=50$ captures most of the measured quality at
the more efficient operating point, while $K_{\mathrm{cand}}=100$ is the
accuracy-oriented configuration used for the main results.

\section{LLM-as-a-Judge Metric Definitions}
\label{appendix:llm-judge}

\begin{table}[H]
\centering
\small
\caption{LLM-as-a-judge evaluation on the three multi-hop benchmarks, scored by Qwen3.7-Plus. All four metrics are percentages (higher is better). Each metric column reports MuSiQue / 2WikiMultiHopQA / HotpotQA, in that order.}
\label{table-llm-judge}
\setlength{\tabcolsep}{2pt}
\begin{tabular*}{\linewidth}{@{\extracolsep{\fill}}lcccc@{}}
\toprule
Method & Answer correctness & Coverage & Context relevancy & Evidence recall \\
\midrule
\multicolumn{5}{c}{\cellcolor{grouphdr}\textit{MuSiQue / 2WikiMultiHopQA / HotpotQA}} \\
GraphRAG & 56.56 / 79.32 / 84.91 & 46.60 / 77.97 / 62.93 & 59.17 / 73.02 / \underline{89.86} & 52.31 / 71.74 / 77.88 \\
LightRAG & 53.07 / 72.52 / 80.98 & 43.54 / 71.15 / 60.88 & 47.55 / 63.65 / 83.57 & 63.91 / 75.89 / 87.49 \\
HyperGraphRAG & \underline{58.03} / 80.31 / 85.00 & \underline{48.52} / 79.73 / \underline{63.78} & \textbf{65.51} / \underline{80.27} / 89.55 & \underline{73.24} / 83.35 / 91.63 \\
HyperRAG & 56.01 / \underline{84.93} / \underline{85.32} & 48.06 / \underline{84.23} / 63.16 & 58.74 / \textbf{80.30} / \textbf{91.13} & 68.23 / 87.90 / 92.37 \\
HippoRAG 2 & 50.09 / 83.93 / 84.91 & 43.20 / 82.42 / 63.31 & 45.75 / 74.58 / 86.65 & 64.77 / \underline{88.45} / \underline{93.65} \\
SAG (Ours) & \textbf{63.69} / \textbf{85.26} / \textbf{87.95} & \textbf{54.73} / \textbf{85.07} / \textbf{64.90} & \underline{59.54} / 76.02 / 89.50 & \textbf{78.87} / \textbf{89.44} / \textbf{95.87} \\
\bottomrule
\end{tabular*}
\end{table}

We follow the LLM-as-a-judge protocol of \citet{xiang2025use} and use Qwen3.7-Plus to evaluate every method. Table~\ref{table-judge-metric-definitions} summarizes the evaluation metrics. Each metric is normalized to a 0--1 scale and multiplied by 100 for presentation in Table~\ref{table-llm-judge}; higher is better.

\begin{table}[H]
\centering
\caption{Definitions of the LLM-as-a-judge evaluation metrics.}
\label{table-judge-metric-definitions}
\scriptsize
\setlength{\tabcolsep}{3pt}
\renewcommand{\arraystretch}{1.15}
\begin{tabularx}{\linewidth}{@{}>{\raggedright\arraybackslash}p{1.65cm}>{\raggedright\arraybackslash}p{2.25cm}>{\raggedright\arraybackslash}X>{\raggedright\arraybackslash}p{3.15cm}>{\centering\arraybackslash}p{0.75cm}@{}}
\toprule
\textbf{Metric} & \textbf{Input} & \textbf{Description} & \textbf{Scoring formula} & \textbf{Output} \\
\midrule
Answer correctness & Generated answer; ground-truth answer; question & Separate calls decompose the ground-truth answers into atomic facts. A third call matches the facts and counts true positives (TP), false positives (FP), and false negatives (FN). & $P=\frac{\mathrm{TP}}{\mathrm{TP}+\mathrm{FP}}$, $R=\frac{\mathrm{TP}}{\mathrm{TP}+\mathrm{FN}}$, $F_1=\frac{2PR}{P+R}$ & $[0,1]$ \\
\addlinespace
Coverage & Generated answer; ground-truth answer; question & Separate calls decompose the generated and ground-truth answers into atomic facts. The judge then determines whether each ground-truth fact is covered by the generated answer. & $N_{\mathrm{covered\ gold}}/N_{\mathrm{gold}}$ & $[0,1]$ \\
\addlinespace
Context relevancy & Retrieved contexts; question & The judge assigns a score of 0 (irrelevant), 1 (partially relevant), or 2 (fully sufficient to answer). Two independent ratings are obtained. & $(s_1+s_2)/4$, where $s_1,s_2\in\{0,1,2\}$ & $[0,1]$ \\
\addlinespace
Evidence recall & Retrieved contexts; gold evidence chunks; question & The judge determines whether each gold evidence item can be inferred from the retrieved contexts. & $N_{\mathrm{entailed}}/N_{\mathrm{gold}}$ & $[0,1]$ \\
\bottomrule
\end{tabularx}
\end{table}

\section{Indexing Prompt for Event and Entity Extraction}
\label{appendix:indexing-prompt}

The offline indexing stage of Section~\ref{subsec:index} issues a single structured-output call per chunk to the indexing LLM (Qwen3.6-Flash in our deployment). The model returns exactly one event together with its typed entities, which are written as one latent hyperedge. Figure~\ref{fig:indexing-prompt} reproduces the system prompt and a representative input/output pair. Note that the example input in the figure includes a \texttt{source\_title} that disagrees with the passage (a fuzzing artifact of the benchmark corpus); the instruction to remain faithful to the passage makes the extractor robust to such noise.

\begin{tcolorbox}[colback=blue!4,colframe=blue!40!black,title=\textbf{System prompt},fonttitle=\small\bfseries,left=4pt,right=4pt,top=3pt,bottom=3pt,boxrule=0.5pt,arc=1.5pt]
\footnotesize
\textbf{Role.} You are a professional content extractor whose core task is to extract two types of structured information from raw documents: \textbf{events} and \textbf{entities}.

\smallskip
\textbf{Event extraction principles.}
\begin{itemize}[leftmargin=1.1em,itemsep=0.5pt,topsep=1pt,parsep=0pt]
\item \emph{Mandatory single event}: regardless of how many different topics the input contains, they must be integrated into one comprehensive event; splitting into multiple events is forbidden.
\item Use short, complete subject--verb--object sentences to state facts; avoid mechanically copying the original text.
\item \emph{Exhaustive information coverage}: all core information from relevant fragments must be reflected in the event content; every sentence must have explicit textual support in the input fragments.
\item \emph{Faithful to the original text}: do not add or omit facts, do not infer on your own, and do not change the original subject.
\item Preserve all key numbers and retain colloquial or relative time expressions in their original wording unless the input explicitly provides an exact date.
\item Ignore advertisements, navigation text, boilerplate, source labels, and fragments without factual value.
\end{itemize}

\smallskip
\textbf{Entity extraction principles.}
\begin{itemize}[leftmargin=1.1em,itemsep=0.5pt,topsep=1pt,parsep=0pt]
\item Extract entities necessary for understanding the event (people, organizations, locations, time, products, key data, etc.), especially subjects and objects.
\item Full names, abbreviations, and aliases of the same entity must all be extracted; entities connected by ``and/or/with'' etc. must be split.
\item Entity types must be selected from \texttt{entity\_types}; the description must clarify the entity's specific role in the event.
\end{itemize}

\smallskip
\textbf{Output requirements.}
\begin{itemize}[leftmargin=1.1em,itemsep=0.5pt,topsep=1pt,parsep=0pt]
\item Return exactly one event containing: \texttt{title} (a short, discriminative English title containing the key subject); \texttt{content} (a self-contained English factual account preserving important subjects, relations, numbers, and time expressions); \texttt{entities} (the entities required to retrieve the event); and \texttt{is\_valid} (\texttt{true} only when the input contains an extractable factual event).
\item For a valid event, \texttt{entities} must contain at least one entity. If no factual event can be extracted, set \texttt{is\_valid} to \texttt{false}, use a short English placeholder title/content, and return an empty \texttt{entities} array.
\item Each entity contains exactly: \texttt{type} (one of the entity types supplied in the input metadata), \texttt{name} (the entity's English name), and \texttt{description} (its specific role or relationship in this event).
\item Return JSON only and follow the supplied response schema exactly: the top-level object contains \texttt{type} set to \texttt{"response"} and \texttt{data}; \texttt{data} contains only an \texttt{items} array holding the single event object.
\item Generate the title, content, entity names, and entity descriptions in English.
\end{itemize}
\end{tcolorbox}

\vspace{0.6em}

\begin{tcolorbox}[colback=orange!6,colframe=orange!60!black,title=\textbf{User message (one request per chunk)},fonttitle=\small\bfseries,left=4pt,right=4pt,top=3pt,bottom=3pt,boxrule=0.5pt,arc=1.5pt]
\footnotesize
The request payload carries the chunk to extract, the closed entity-type inventory, and a short previous-chunk context:
\begin{itemize}[leftmargin=1.1em,itemsep=0.5pt,topsep=1pt,parsep=0pt]
\item \texttt{items[0].idx}: \texttt{26}; \texttt{items[0].title}: \texttt{Clement Attlee}; \texttt{items[0].text}: the raw passage associated with this item:
\begin{lstlisting}[style=rawtext]
Clement Richard Attlee, 1st Earl Attlee, {'1': ", '2': ", '3': ", '4': "} (3 January 1883 -- 8 October 1967) was a British Labour politician who served as Prime Minister of the United Kingdom from 1945 to 1951 and Leader of the Labour Party from 1935 to 1955. In 1940, Attlee took Labour into the wartime coalition government and served under Winston Churchill, becoming the first person to hold the office of Deputy Prime Minister of the United Kingdom. He went on to lead the Labour Party to an unexpected landslide victory at the 1945 general election; forming the first Labour majority government, and a mandate to implement its postwar reforms. The 12.0\% national swing from the Conservatives to Labour was unprecedented at that time and remains the largest ever achieved by any party at a general election in British political history. He was re-elected with a narrow majority at the 1950 general election. In the following year, Attlee called a snap general election, hoping to increase his parliamentary majority. However, he was narrowly defeated by the Conservatives under the leadership of Winston Churchill; despite winning the most votes of any political party in any general election in British political history until the Conservative Party's fourth consecutive victory in 1992. Attlee remains the longest-ever serving Leader of the Labour Party.
\end{lstlisting}
\item \texttt{meta.entity\_types}: the 11 closed types of Section~\ref{subsec:index} (\texttt{time}, \texttt{location}, \texttt{person}, \texttt{organization}, \texttt{group}, \texttt{subject}, \texttt{work}, \texttt{product}, \texttt{action}, \texttt{metric}, \texttt{tags}), each with a short definition and examples.
\item \texttt{meta.source\_type}/\texttt{source\_title}/\texttt{source\_summary}: document-level metadata (not necessarily aligned with the passage).
\item \texttt{meta.previous\_context}: the head of the previous chunk, used to resolve cross-chunk references.
\end{itemize}
\end{tcolorbox}

\begin{tcolorbox}[colback=green!5,colframe=green!50!black,title=\textbf{Example output (full model response)},fonttitle=\small\bfseries,left=4pt,right=4pt,top=3pt,bottom=3pt,boxrule=0.5pt,arc=1.5pt,breakable,enhanced]
The complete, unabridged JSON returned by the model for the example chunk (all 15 entities shown):
\begin{lstlisting}[style=jsonout]
{
  "type": "response",
  "data": {
    "items": [
      {
        "title": "Clement Attlee",
        "content": "Clement Richard Attlee, 1st Earl Attlee, {'1': \", '2': \", '3': \", '4': \"} (3 January 1883 -- 8 October 1967) was a British Labour politician who served as Prime Minister of the United Kingdom from 1945 to 1951 and Leader of the Labour Party from 1935 to 1955. In 1940, Attlee took Labour into the wartime coalition government and served under Winston Churchill, becoming the first person to hold the office of Deputy Prime Minister of the United Kingdom. He went on to lead the Labour Party to an unexpected landslide victory at the 1945 general election; forming the first Labour majority government, and a mandate to implement its postwar reforms. The 12.0% national swing from the Conservatives to Labour was unprecedented at that time and remains the largest ever achieved by any party at a general election in British political history. He was re-elected with a narrow majority at the 1950 general election. In the following year, Attlee called a snap general election, hoping to increase his parliamentary majority. However, he was narrowly defeated by the Conservatives under the leadership of Winston Churchill; despite winning the most votes of any political party in any general election in British political history until the Conservative Party's fourth consecutive victory in 1992. Attlee remains the longest-ever serving Leader of the Labour Party.",
        "entities": [
          {"type": "person", "name": "Clement Attlee", "description": "British Labour politician who served as Prime Minister from 1945 to 1951 and Leader of the Labour Party from 1935 to 1955."},
          {"type": "time", "name": "1945 to 1951", "description": "The period during which Clement Attlee served as Prime Minister of the United Kingdom."},
          {"type": "time", "name": "1935 to 1955", "description": "The period during which Clement Attlee served as Leader of the Labour Party."},
          {"type": "time", "name": "1940", "description": "The year Attlee joined the wartime coalition government and became Deputy Prime Minister."},
          {"type": "time", "name": "1945", "description": "The year of the general election where Attlee led Labour to a landslide victory."},
          {"type": "time", "name": "1950", "description": "The year of the general election where Attlee was re-elected with a narrow majority."},
          {"type": "time", "name": "1951", "description": "The year Attlee called a snap general election and was narrowly defeated."},
          {"type": "organization", "name": "Labour Party", "description": "The political party led by Attlee from 1935 to 1955 and which he led to victory in 1945."},
          {"type": "location", "name": "United Kingdom", "description": "The country where Attlee served as Prime Minister and where the general elections took place."},
          {"type": "person", "name": "Winston Churchill", "description": "Prime Minister under whom Attlee served as Deputy Prime Minister in 1940, and the Conservative leader who defeated Attlee in 1951."},
          {"type": "metric", "name": "12.0%", "description": "The unprecedented national swing from the Conservatives to Labour in the 1945 general election."},
          {"type": "subject", "name": "Deputy Prime Minister of the United Kingdom", "description": "The office Attlee held starting in 1940, being the first person to hold it."},
          {"type": "subject", "name": "1945 general election", "description": "The election where Attlee led Labour to a landslide victory and the largest swing in British history."},
          {"type": "subject", "name": "1950 general election", "description": "The election where Attlee was re-elected with a narrow majority."},
          {"type": "subject", "name": "snap general election", "description": "The election called by Attlee in 1951, which resulted in his narrow defeat."}
        ],
        "is_valid": true
      }
    ]
  }
}
\end{lstlisting}
\end{tcolorbox}
\begin{minipage}{\linewidth}
\captionof{figure}{Prompt used for offline event--entity extraction. The system prompt enforces one comprehensive event per chunk with exhaustive, faithful coverage; the user message supplies the chunk, the closed entity-type inventory, and previous-chunk context; the model returns the event and typed entities that form one latent hyperedge.}
\label{fig:indexing-prompt}
\end{minipage}

\section{LLM Reranking Prompt Input and Output}
\label{appendix:reranking-prompt}

The LLM reranking stage receives a multi-hop question and candidate relationship descriptions and returns up to five useful relation indices. Figure~\ref{fig:reranking-prompt} reproduces the prompt exchange in the same compact visual style as Figure~\ref{fig:indexing-prompt}. The candidate descriptions are represented by their source indices, while the remaining prompt and response content is retained.

\begin{tcolorbox}[colback=blue!4,colframe=blue!40!black,title=\textbf{System prompt},fonttitle=\small\bfseries,left=4pt,right=4pt,top=3pt,bottom=3pt,boxrule=0.5pt,arc=1.5pt]
\footnotesize
I will provide you with a set of relationship descriptions from a knowledge graph. Select up to 5 relationships most useful for answering this multi-hop question.

\smallskip
Return JSON with \texttt{thought\_process} and \texttt{useful\_relations} (list of up to 5 relation indices like \texttt{[id]}, most useful first).
\end{tcolorbox}

\vspace{0.6em}

\begin{tcolorbox}[colback=orange!6,colframe=orange!60!black,title=\textbf{User message (target request)},fonttitle=\small\bfseries,left=4pt,right=4pt,top=3pt,bottom=3pt,boxrule=0.5pt,arc=1.5pt]
\footnotesize
\textbf{Question:} What month did the Tripartite discussions begin between Britain, France, and the country where, despite being headquartered in the nation called the nobilities commonwealth, the top-ranking Warsaw Pact operatives originated?

\smallskip
\textbf{Relationship descriptions:} \texttt{[0], [1], [2], ..., [99]}
\end{tcolorbox}

\vspace{0.6em}

\begin{tcolorbox}[colback=green!5,colframe=green!50!black,title=\textbf{Assistant message (target output)},fonttitle=\small\bfseries,left=4pt,right=3pt,top=3pt,bottom=3pt,boxrule=0.5pt,arc=1.5pt,breakable,enhanced]
\begin{lstlisting}[style=jsonout]
  "thought_process": "1. Identify the country where the 'top-ranking Warsaw Pact operatives originated' and which is 'headquartered in the nation called the nobilities commonwealth'. From relation [43], the 'Polish-Lithuanian Commonwealth' is referred to as the 'nobilities commonwealth'. The capital of the Warsaw Pact's Combined Command was in Warsaw, Poland (relation [3]). Thus, the country in question is Poland.\n2. Identify the 'Tripartite discussions' involving Britain, France, and Poland. Relation [22] describes 'Tripartite negotiations' involving the USSR, Britain, and France, but the question specifies Britain, France, and Poland. However, historical context often links these negotiations to the broader Allied talks. Let's re-read carefully. The question asks about discussions between Britain, France, and the country identified (Poland). Wait, relation [22] says 'mid-June, the main Tripartite negotiations started' and mentions 'Britain opposed such proposals...'. It doesn't explicitly list Poland as a party to the start of the negotiations in the text, but the historical event is the Anglo-Franco-Soviet talks. Let's look for a relation that mentions the start date. Relation [22] states: 'By the end of May, drafts were formally presented, and in mid-June, the main Tripartite negotiations started.'\n3. Determine the month. Relation [22] says 'mid-June'. So the month is June.\n4. Select the most useful relations to confirm the entities and the date.\n   - [43] identifies 'Poland' as the 'nobilities commonwealth'.\n   - [3] identifies 'Poland' (Warsaw) as the headquarters of the Warsaw Pact operatives, confirming the country is Poland.\n   - [22] provides the date 'mid-June' for the start of the 'Tripartite negotiations' involving the relevant powers (Britain, France, USSR - though the question says Britain, France, and the country... actually, the question phrasing 'between Britain, France, and the country...' might be a trick or referring to a specific bilateral/multilateral context. However, given the options, [22] is the only one describing 'Tripartite discussions' with a date. The 'Tripartite' usually refers to UK, France, USSR. Does the question imply Poland was a party? Or is it referring to the negotiations regarding Poland? 'The discussion focused on potential guarantees to central and east European countries...'. The 'Tripartite' usually means the three great powers. Let's assume the question refers to the events in [22].\n   - Let's check if there are other 'Tripartite' mentions. [93] mentions 'Tripartite Pact' (Germany, Italy, Japan). Not relevant.\n   - So the key is linking 'nobilities commonwealth' -> Poland -> Warsaw Pact HQ -> Poland. Then finding the Tripartite talks involving Britain and France (and implicitly the USSR, or perhaps the question implies the talks about the country?). The text in [22] says 'Tripartite negotiations started... mid-June'.\n   - Therefore, the relations needed are: [43] (Poland = nobilities commonwealth), [3] (Warsaw Pact HQ in Poland, confirming the origin of operatives), and [22] (Date of Tripartite negotiations). We need 5 relations.\n   - [1] also mentions Warsaw Pact established in Warsaw, Poland. Useful to reinforce [3].\n   - [12] mentions Solidarity in Poland, confirming Poland's identity in the Warsaw Pact context.\n   - So the set is [43], [3], [22], [1], [12].",
  "useful_relations": ["[43]", "[3]", "[22]", "[1]", "[12]"]
\end{lstlisting}
\end{tcolorbox}

\begin{minipage}{\linewidth}

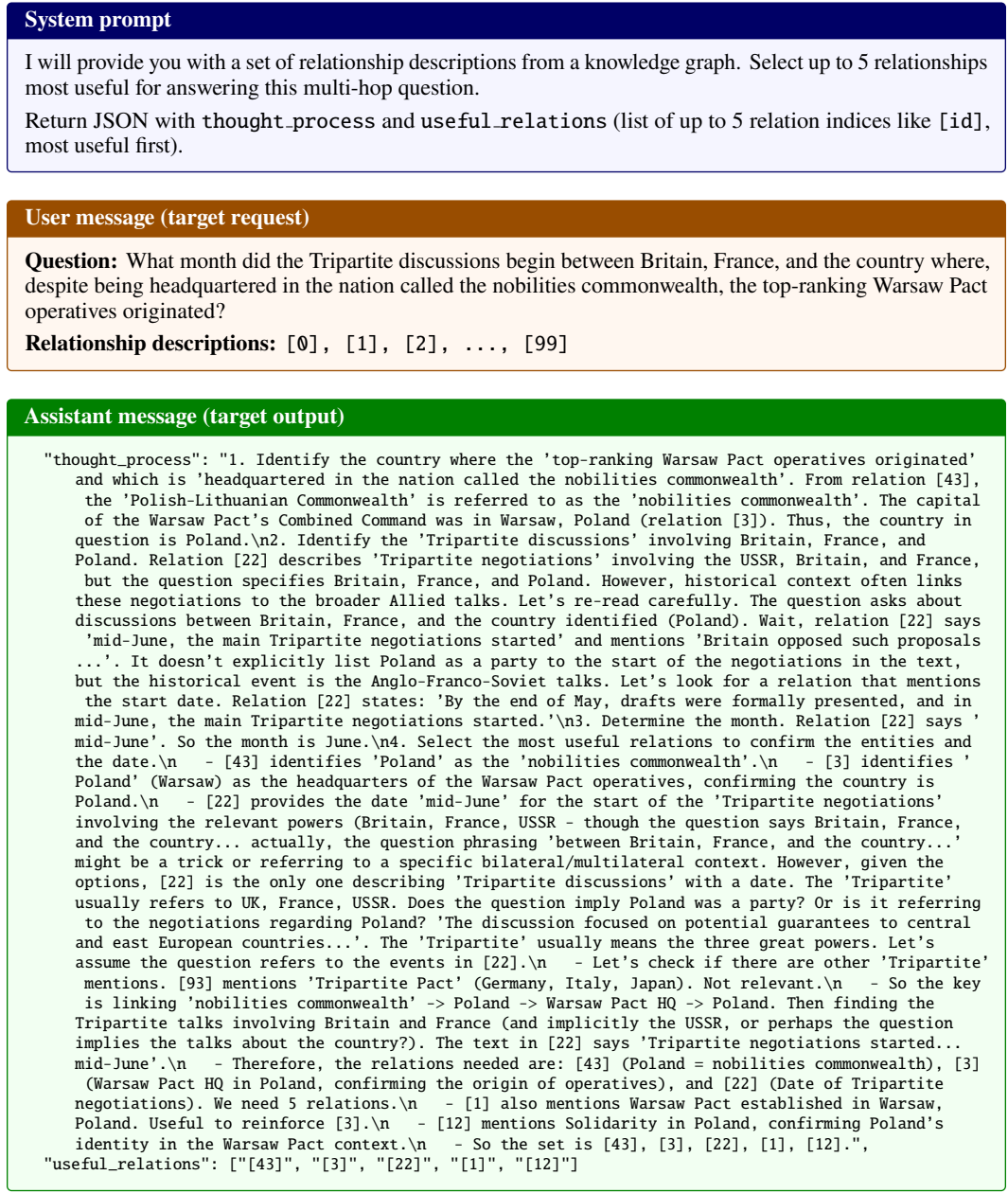
\captionof{figure}{LLM reranking prompt exchange; candidate descriptions are identified by their source indices, while the remaining fields are shown in full.}
\label{fig:reranking-prompt}
\end{minipage}

\section{Results on NarrativeQA}
\label{appendix:narrativeqa}

The three primary benchmarks of Section~\ref{sec:mainresults} all evaluate
multi-hop evidence chaining over encyclopedic passages. To probe whether SAG's
event--entity organization also benefits workloads that hinge on semantic
comprehension rather than strict multi-hop fact chaining, we additionally
evaluate on NarrativeQA \citep{kocisky2018narrativeqa}, a
reading-comprehension benchmark built from books and movie scripts whose
questions require tracking characters, temporal ordering, and narrative
context rather than composing evidence across independent documents.
Following the unified configuration of
Section~\ref{subsec:baselines} (BGE-Large-EN-v1.5 retriever, Qwen3.6-Flash
reader), we sample 293 development questions and pool all supporting and
distractor passages into a retrieval corpus of 4,111 passages, consistent with
the setup of the primary benchmarks.
Table~\ref{table-narrativeqa} compares SAG against HippoRAG~2, the
strongest structure-augmented baseline in the main results.

\begin{table}[H]
\centering
\small
\caption{QA results (EM/F1) on NarrativeQA under the unified configuration
(BGE-Large-EN-v1.5 retriever, Qwen3.6-Flash reader). \textbf{Bold} marks the
better value.}
\label{table-narrativeqa}
\setlength{\tabcolsep}{8pt}
\begin{tabular}{lcc}
\toprule
Method & EM & F1 \\
\midrule
HippoRAG~2 & 6.48 & 22.86 \\
SAG (Ours) & \textbf{12.29} & \textbf{31.90} \\
\bottomrule
\end{tabular}
\end{table}

SAG nearly doubles Exact Match (6.48\% to 12.29\%) and raises F1 from 22.86\%
to 31.90\% over HippoRAG~2. NarrativeQA scoring is stringent: exact-match
answers are required for EM credit and answers are often short spans, so the
gain indicates that SAG retrieves the precise evidence needed to answer.
This is consistent with SAG's mechanism: each chunk is organized into a
semantically complete event with typed indexing entities, and query-time
shared-entity joins link evidence that must be assembled across a long
narrative, whereas HippoRAG~2 propagates embedding-derived relevance through
personalized PageRank whose damping attenuates remote associations. Together
with the multi-hop results, this suggests SAG's event--entity organization
transfers beyond fact-chaining to narrative comprehension.

\section{QA Input Context Formats}
\label{appendix:qa-context}

In the QA stage, each method assembles the retrieved evidence into a context
that the reader consumes. The composition of this context differs
substantially across methods and directly shapes how the QA results of
Section~\ref{sec:mainresults} should be interpreted.
Table~\ref{table-qa-context} summarizes the QA input context of each method.

\begin{table}[H]
\centering
\small
\caption{Composition of the QA input context for each evaluated method.
``Passage'' refers to the raw retrieved chunk text. GraphRAG concatenates five
segments (reports, entities, relationships, optional claims, and sources);
HyperGraphRAG, HyperRAG (hypergraph mode), and LightRAG format retrieved
evidence as three-segment CSV rows; HippoRAG~2 and SAG feed the raw passage
text directly to the reader.}
\label{table-qa-context}
\setlength{\tabcolsep}{5pt}
\begin{tabular}{lp{0.66\linewidth}}
\toprule
Method & QA context composition \\
\midrule
GraphRAG & Reports + Entities + Relationships + (Claims) + Sources, five segments concatenated \\
HippoRAG~2 & Raw retrieved passage full text (top 5) \\
HyperGraphRAG & Entities + Relationships + Sources, three-segment CSV \\
HyperRAG & naive mode: raw chunk text; hypergraph mode: Entities + Relationships + Sources, three-segment CSV \\
LightRAG & Entities + Relationships + Sources, three-segment CSV \\
SAG (Ours) & Raw retrieved passage full text (top 5) \\
\bottomrule
\end{tabular}
\end{table}

Two points follow from this comparison. First, SAG's QA gains in
Section~\ref{sec:mainresults} and Appendix~\ref{appendix:narrativeqa} are not
attributable to a richer reader context: SAG supplies the reader with exactly
the same input modality as HippoRAG~2---raw passage text---yet achieves higher
EM and F1, so the improvement must originate from the retrieval stage rather
than from any difference in context formatting. Second, methods that structure
the reader input (GraphRAG, HyperGraphRAG, HyperRAG, LightRAG) provide entity
and relationship summaries that can partially answer a question even when the
underlying evidence is imperfectly retrieved, so their QA numbers and
LLM-judge context-relevancy scores are not directly comparable to methods that
rely solely on raw passages; this should be kept in mind when reading
Table~\ref{table-llm-judge}.

\end{document}